\PassOptionsToPackage{table,dvipsnames}{xcolor}

\documentclass[10pt,twocolumn,letterpaper]{article}

\usepackage[pagenumbers]{iccv} 

\usepackage{multirow}
\usepackage{listings}
\usepackage{placeins}

\definecolor{iccvblue}{rgb}{0.21,0.49,0.74}
\usepackage[pagebackref,breaklinks,colorlinks,allcolors=iccvblue]{hyperref}

\def\paperID{*****} 
\def\confName{ICCV}
\def\confYear{2025}

\title{TalkVid: A Large-Scale Diversified Dataset for Audio-Driven Talking Head Synthesis}

\author{Shunian Chen$^{1,*}$, 
Hejin Huang$^{1,2,*}$, 
Yexin Liu$^{3,}$\thanks{First three authors contributed to this work equally. Ser-Nam Lim, Harry Yang, and Benyou Wang are the corresponding authors.}, 
Zihan Ye$^{1}$,
Pengcheng Chen$^{1}$, \\ 
Chenghao Zhu$^{1}$, 
Michael Guan$^{1}$,
Rongsheng Wang$^{1}$, 
Junying Chen$^{1}$,
Guanbin Li$^{2}$, \\
Ser-Nam Lim$^{3,\dag}$,
Harry Yang$^{3,\dag}$,
Benyou Wang$^{1,\dag}$ \\
$^{1}$The Chinese University of Hong Kong, Shenzhen \\ $^{2}$Sun Yat-sen University \\
$^{3}$The Hong Kong University of Science and Technology \\
\texttt{wangbenyou@cuhk.edu.cn}
}

\begin{document}
\maketitle
\begin{abstract}
Audio-driven talking head synthesis has achieved remarkable photorealism, yet state-of-the-art (SOTA) models exhibit a critical failure: they lack generalization to the full spectrum of human diversity in ethnicity, language, and age-groups. We argue this generalization gap is a direct symptom of limitations in existing training data, which lack the necessary scale, quality, and diversity. To address this challenge, we introduce \texttt{TalkVid}, a new large-scale, high-quality, and diverse dataset containing 1244 hours of video from 7729 unique speakers. \texttt{TalkVid} is curated through a principled, multi-stage automated pipeline that rigorously filters for motion stability, aesthetic quality, and facial detail, and is validated against human judgments to ensure its reliability. Furthermore, we construct and release \texttt{TalkVid-Bench}, a stratified evaluation set of 500 clips meticulously balanced across key demographic and linguistic axes. Our experiments demonstrate that a model trained on \texttt{TalkVid} outperforms counterparts trained on previous datasets, exhibiting superior cross-dataset generalization. Crucially, our analysis on \texttt{TalkVid-Bench} reveals performance disparities across subgroups that are obscured by traditional aggregate metrics, underscoring its necessity for future research. Code and data can be found in \url{https://github.com/FreedomIntelligence/TalkVid}
\end{abstract}    
\section{Introduction}
\label{sec:introduction}

\begin{figure*}[t]
    \centering
    \includegraphics[width=\linewidth]{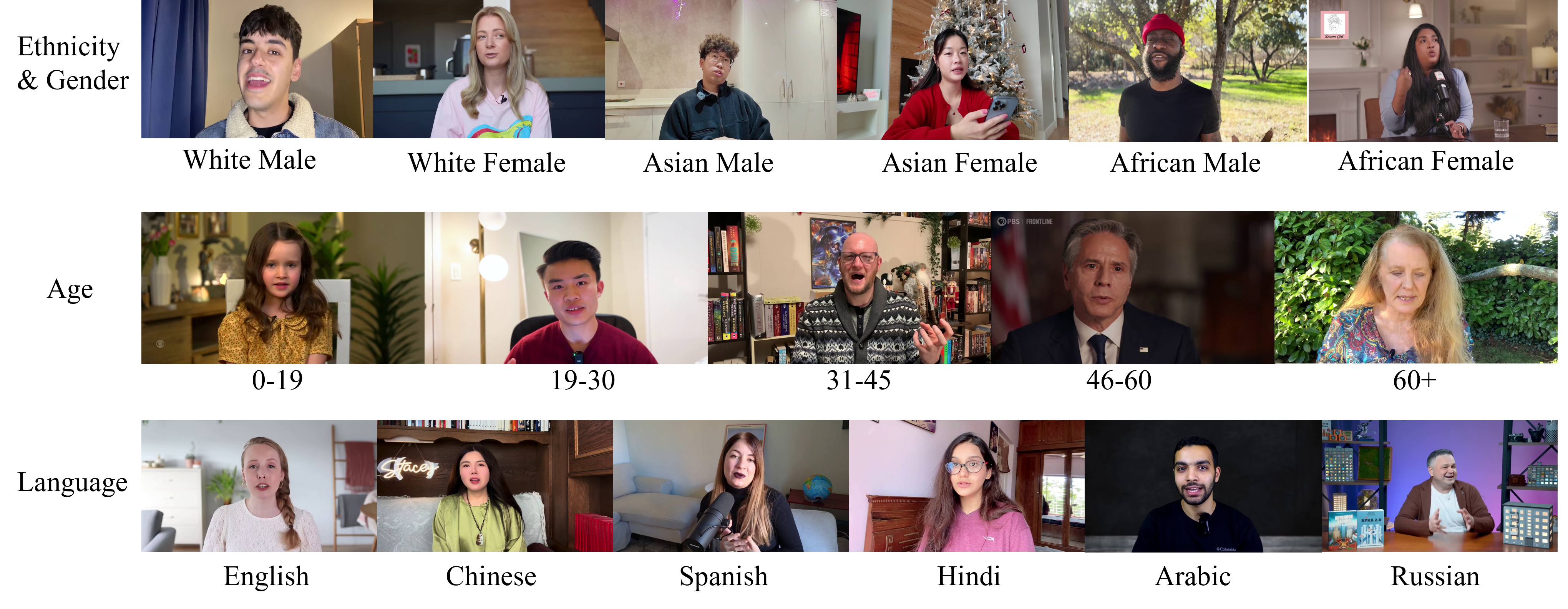} 
    \caption{Examples from our \texttt{TalkVid} dataset, showcasing the diversity in identity, ethnicity, and head pose that SOTA models must generalize to. Existing datasets lack this combined diversity and technical quality, leading to generalization failures.}
    \label{fig:teaser}
\end{figure*}

Audio-driven talking head synthesis has achieved remarkable photorealism, with recent models generating outputs that are often indistinguishable from real video under controlled conditions~\citep{cui2024hallo3, xu2024vasa}. Yet, this success masks a critical fragility. SOTA models remain brittle; as our experiments will demonstrate, their performance degrades significantly when confronted with the full spectrum of human diversity. This generalization gap—the failure to handle varied ethnicities, unconstrained head poses, and diverse languages as shown in Figure~\ref{fig:teaser}—is not a minor flaw. It is the primary bottleneck preventing the widespread, reliable, and equitable application of this technology.

We argue this brittleness is a direct consequence of a foundational bottleneck: the data upon which these models are trained. Existing datasets force a difficult trade-off. On one hand, datasets like HDTF~\cite{zhang2021flow} offer high-resolution, front-facing videos but are narrowly curated, lacking the linguistic, demographic, and motion diversity needed for robust generalization. On the other hand, large-scale, ``in-the-wild" datasets like VoxCeleb2~\cite{chung2018voxceleb2} offer diversity but are rife with technical artifacts—motion blur, compression noise, and inconsistent framing—that compromise the training of high-fidelity generative models. Consequently, the field has lacked a resource that is simultaneously \textbf{large-scale, diverse, and technically pristine}. This data gap directly translates into models that are biased and unreliable.

\begin{table*}[t]
\centering
\scriptsize

\setlength{\tabcolsep}{5pt}
\begin{tabular}{lccccccccc}

\toprule
\textbf{Name} & \textbf{Year}  & \textbf{Speaker} & \textbf{Hours} & \textbf{Resolution} & \textbf{Language} & \textbf{Age}   & \textbf{Body Included}  & \textbf{Caption} & \textbf{Source} \\
\midrule
GRID~\cite{cooke2006audio}         & 2006    & 33   & 27.5        & 288p, 576p &  English & 18–49 & No & No  & Lab  \\
Crema-D~\cite{cao2014crema}        & 2014    & 91   & 11.1         & 720p               & English & 20–74 &No & No  & Lab  \\
LRW~\cite{chung2017lip}            & 2017    & 1k+  & 173       & 120p               & English & -  &No & No  & Wild \\
VoxCeleb1~\cite{nagrani2020voxceleb} & 2017  & 1.2k  & 352    & 256p               &- &- &No & No  & Wild \\
VoxCeleb2~\cite{chung2018voxceleb2} & 2018   & 6.1k+   &\textbf{2.4k}     & 256p               &- &- &No & No  & Wild \\
MEAD~\cite{wang2020mead}           & 2020    & 60    & 39       & 384p                &  English& 20–35& No & No  & Lab  \\
HDTF~\cite{zhang2021flow}          & 2021    & 362   & 15.8       & 512p                & - & - & No & No  & Wild \\
Hallo3~\cite{cui2024hallo3}        & 2024    & -    & 70          & 480p               & - & - & Yes & No  & Wild \\
MultiTalk~\cite{sung2024multitalk} & 2024    & -    & 243.2        & 512p                & \textbf{20 lang} & - & Yes & No  & Wild \\
\rowcolor[gray]{0.9}
\textbf{TalkVid (Ours)} & 2025  & \textbf{7,729}  & 1,244.33 & 1080p,2160p & 15 lang & 0-60+ & Yes & \textbf{Yes} & Wild \\
\bottomrule
\end{tabular}

\caption{Comparison of open-source datasets for audio-driven talking-head generation.}

\begin{flushleft}
\scriptsize
Resolution abbreviations denote pixel dimensions (width$\times$height): 120p (120$\times$120), 256p (256$\times$256), 288p (360$\times$288), 384p (384$\times$384), 480p (480$\times$720), 512p (512$\times$512), 576p (720$\times$576), 720p (960$\times$720), 1080p (1920$\times$1080), and 2160p (3840$\times$2160). For our \texttt{TalkVid} dataset, the listed resolutions comprise 87.59\% of the data.
\end{flushleft}
\label{tab:datasets_statistics}
\end{table*}

This paper introduces a unified solution to this data-centric challenge. We present \texttt{TalkVid}, a new dataset designed from the ground up to eliminate the trade-offs of prior work. Its construction is guided by three core principles: \textbf{1) Scale and Diversity}, sourcing over 6,000 hours of raw video to capture a broad cross-section of speakers, languages, and contexts; \textbf{2) High-Quality}, enforced via a rigorous, multi-stage automated filtering pipeline that ensures technical excellence in motion, aesthetics, and facial detail; and \textbf{3) Reliability}, validated through human verifications confirming our pipeline's alignment with perceptual quality. 

Crucially, better training requires better evaluation. We introduce \texttt{TalkVid-Bench}, a dedicated 500-clip evaluation benchmark stratified across key demographic (age, gender, ethnicity) and linguistic dimensions. Current evaluation practices, which rely on aggregate metrics, obscure critical failure modes and fairness issues. \texttt{TalkVid-Bench} enables a fine-grained analysis that reveals model-specific biases and quantifies true generalization. Our experiments, leveraging this benchmark, provide the definitive evidence that training on \texttt{TalkVid} yields superior performance and that \texttt{TalkVid-Bench} is essential for exposing robustness gaps invisible to previous evaluation protocols.

Our \textbf{contributions} are threefold: 1) we introduce \texttt{TalkVid}, a large-scale, high-quality dataset containing 1,244 hours of talking-head videos from 7,729 speakers, curated via a rigorous, human-validated pipeline;
2) we build \texttt{TalkVid-Bench}, a stratified evaluation benchmark balanced across demographic and linguistic dimensions to enable transparent assessment of model fairness and generalization;
3) we conduct comprehensive empirical validation demonstrating that models trained on \texttt{TalkVid} achieve state-of-the-art performance, superior cross-domain robustness, and reveal critical biases missed by existing methods, thereby providing the community with essential resources for developing equitable talking-head synthesis models.

\begin{figure*}[htbp]
    \centering
    \includegraphics[width=\linewidth]{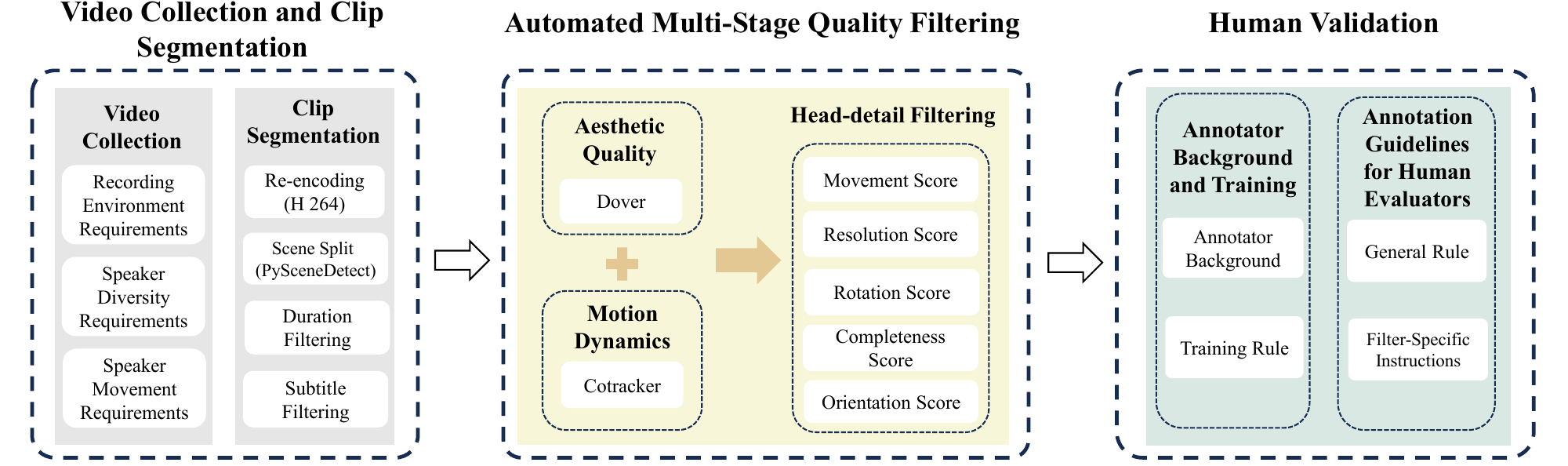}
    \caption{The \texttt{TalkVid} construction pipeline. The process starts with (1) video collection and clip segmentation. Each candidate clip then undergoes (2) a multi-stage filtering cascade to enforce quality across aesthetics, motion, and facial detail. Finally, the pipeline's effectiveness is (3) validated against human judgments.}
    \label{fig:construction_pipeline}
\end{figure*}
\section{Related Works}

\paragraph{Audio-Driven Talking Head Synthesis.}
The synthesis of talking heads from audio has rapidly evolved, moving from GAN-based architectures to the now-dominant diffusion models. Early methods, often leveraging Generative Adversarial Networks (GANs)~\cite{karras2019style}, achieved high-resolution, efficient synthesis. For instance, StyleHEAT~\cite{yin2022styleheat} enabled one-shot generation, while others like SadTalker~\cite{zhang2023sadtalker} and LipSync3D~\cite{prajwal2020lip} focused on improving motion dynamics and lip-sync accuracy, respectively. Despite these advances, GAN-based approaches often suffer from temporal inconsistencies and limited expressiveness, particularly with large pose variations.

To address these limitations, recent work has overwhelmingly shifted towards Diffusion Models (DMs), which offer superior temporal stability and photorealism. Foundational models like VExpress~\cite{wang2024v} demonstrated the potential of DMs for this task. Subsequent works have focused on enhancing control and alignment; for example, AniPortrait~\cite{wei2024aniportrait} and Hallo~\cite{xu2024hallo} employ multi-stage pipelines and part-aware modules to improve audio-visual correspondence. The current SOTA, exemplified by VASA~\cite{xu2024vasa}, Hallo3~\cite{cui2024hallo3}, and EDTalk~\cite{tan2024edtalk}, introduces disentangled latent representations to generate highly expressive and controllable facial dynamics beyond simple lip movements. However, these methods often rely on complex pipelines and pre-trained priors, leaving a gap for a unified and efficiently controllable model.

\paragraph{Datasets for Talking Head Generation.}
Progress in talking head synthesis is intrinsically tied to the available training data. Early datasets, such as GRID~\cite{cooke2006audio} and CREMA-D~\cite{cao2014crema}, were collected in controlled laboratory settings, providing clean audio-visual pairs but lacking diversity and scale. The advent of large-scale, "in-the-wild" datasets like LRW~\cite{chung2017lip} and VoxCeleb2~\cite{chung2018voxceleb2} was a major step forward, offering real-world variability. However, these datasets often suffer from inconsistent video quality and a lack of granular annotations necessary for modeling fine-grained expression. More recent high-quality datasets, including MEAD~\cite{wang2020mead}, HDTF~\cite{zhang2021flow}, and MultiTalk~\cite{sung2024multitalk}, have improved upon resolution and speaker diversity. Nevertheless, a critical bottleneck persists: the absence of a truly large-scale benchmark that pairs high-resolution video with rich, semantic annotations for detailed expressive and semantic control. Our dataset is designed to fill this void, providing over 1,244 hours of high-fidelity video with dense captions to foster the next generation of controllable talking head models.

\begin{figure*}[t]
    \centering
    \includegraphics[width=\textwidth]{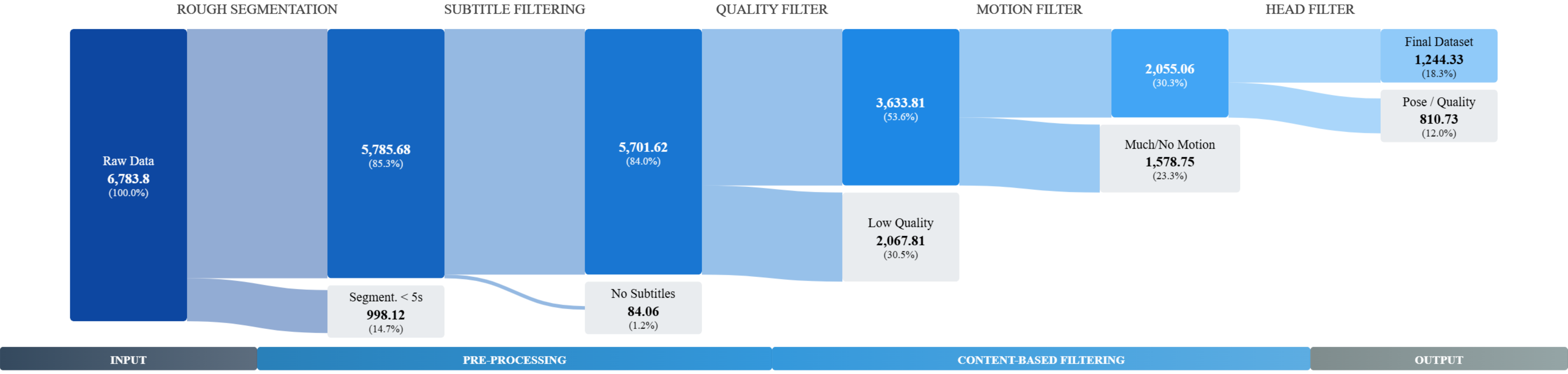}
    \caption{The data filtering cascade. This diagram quantifies the progressive refinement of the dataset, showing the hours of video retained and discarded at each preprocessing and content-based filtering stage.}
    \label{fig:sankey_plot_for_data_filter}
\end{figure*}

\begin{figure*}[t]
    \centering
    \includegraphics[width=\textwidth]{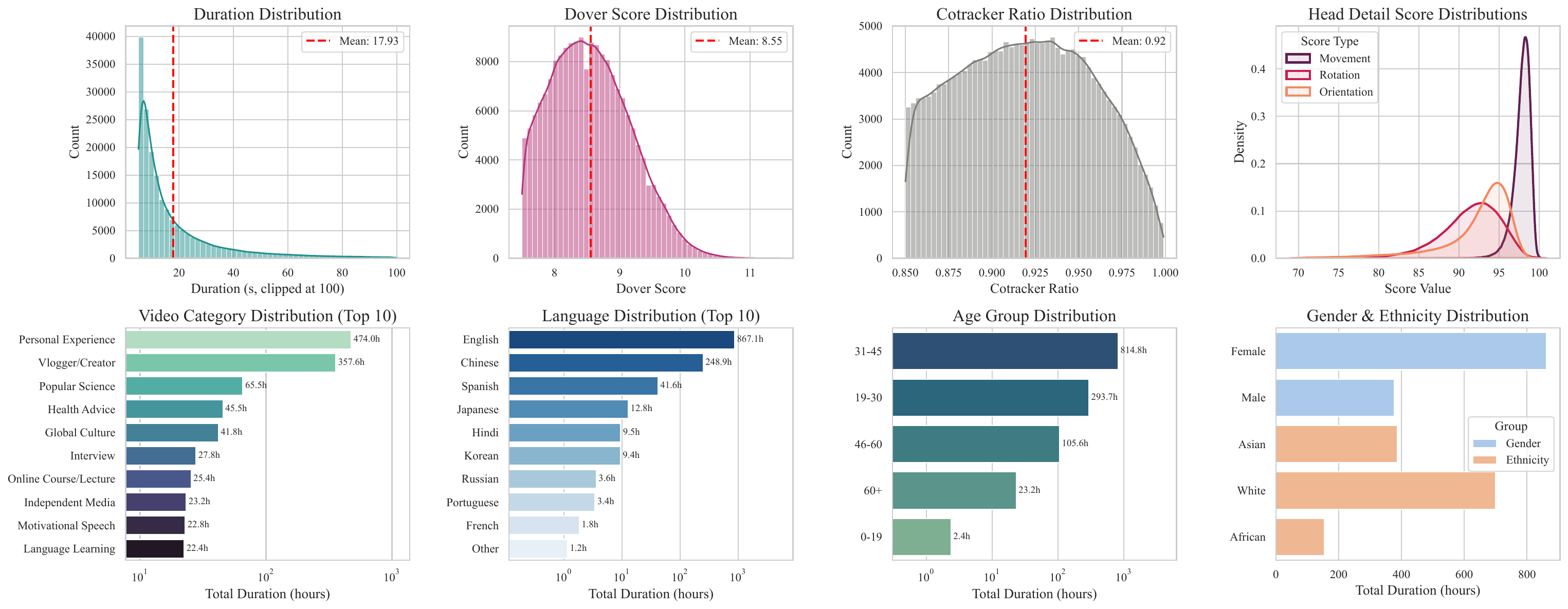}
    \caption{Statistical distributions of the TalkVid dataset. Top row: technical quality metrics for the final, filtered dataset. Bottom row: distributions of the high-level characteristics, including video categories, language, and speaker demographics.}
    \label{fig:data_distribution}
\end{figure*}
\section{The TalkVid Dataset}
\label{sec:dataset}
This section details the construction methodology, human verification protocol, and presents the analysis of the dataset's characteristics.

\subsection{Construction Methodology}

The TalkVid construction pipeline is illustrated in Figure~\ref{fig:construction_pipeline}. We first describe each stage and then present a quantitative analysis validating the effectiveness of our filtering process against human assessments.

\subsubsection{Data Preprocess}

\paragraph{Stage 1: Video Collection.} We begin by collecting over 30,000 videos from YouTube, totaling more than 6,000 hours of high-resolution (1080p or higher) content. To ensure a high-quality starting point, we target genres known for clear audio and stable video, such as educational lectures, technical reviews, and professional vlogs. For each source, we download the video, audio, and available auto-generated transcripts through \texttt{yt-dlp}~\cite{yt-dlp-github}. 
Complete sourcing criteria can be found in Appendix~\ref{app:video_collection_criteria}.

\paragraph{Stage 2: Clip Segmentation.} The collected videos are then processed through a segmentation pipeline. First, all videos are standardized by re-encoding them to the H.264 (MP4) format. We then use \texttt{PySceneDetect}~\cite{PySceneDetect} to detect shot boundaries. Segments shorter than 5 seconds are discarded, as they are typically too brief to contain a complete thought or gesture. Finally, using transcript timings, we remove segments without speech events.

\subsubsection{Content-Based Filtering}
To ensure each clip meets the technical demands of modern generative modeling, we subject each candidate to a content-based filtering cascade. A clip is retained only if it satisfies all criteria across three key quality dimensions: aesthetic quality, motion dynamics, and head detail. The metrics and their corresponding thresholds are summarized in Table~\ref{tab:filtering_criteria}, while the rationale for each is elaborated below.

\begin{table}[t]
\scriptsize
\centering
\begin{tabular}{@{}lll@{}}
\toprule
\textbf{Filtering Stage} & \textbf{Metric} & \textbf{Criterion / Threshold} \\
\midrule
Aesthetic Quality & DOVER Score~\cite{wu2023exploring} & $\geq 7.0$ \\
\midrule
Motion Stability & CoTracker Ratio~\cite{karaev2024cotracker3} & $\in [0.85, 0.999]$ \\
\midrule
\multirow{5}{*}{Head Detail} & Movement Score & Avg $\geq 80$, Min $\geq 60$ \\
 & Rotation Score & Avg $\geq 70$, Min $\geq 60$ \\
 & Orientation Score & Avg $\geq 70$, Min $\geq 30$ \\
 & Resolution Score & Avg $\geq 50$, Min $\geq 40$ \\
 & Completeness Score & $= 100$ \\
\bottomrule
\end{tabular}
\caption{Filtering criteria for video clip selection, each candidate clip must satisfy all listed conditions. }
\label{tab:filtering_criteria}
\end{table}

\paragraph{Filter 1: Aesthetic Quality.}
To guarantee high visual fidelity, we employ DOVER~\cite{wu2023exploring}, a no-reference video quality assessment model. By enforcing a minimum score, we filter out clips containing perceptible compression artifacts, noise, or excessive blur, retaining only those with a clean, high-quality appearance.

\paragraph{Filter 2: Motion Dynamics.}
We select for clips with natural motion characteristics using the point tracking stability ratio from CoTracker~\cite{karaev2024cotracker3}. The specified range serves a dual purpose: the lower bound ($\geq 0.85$) removes clips with erratic motion or blur, which manifest as tracking failures, while the upper bound ($\leq 0.999$) crucially discards unnaturally static or "frozen" shots. This ensures the retention of subtle, natural movements characteristic of a live speaker.

\paragraph{Filter 3: Head-Detail Filtering.}
Finally, we perform a fine-grained assessment of the subject's head using a suite of five metrics designed to ensure stability and clarity. These metrics collectively ensure the temporal stability of facial keypoints (Movement Score), smoothness of head orientation transitions (Rotation Score), a largely frontal view without extreme angles (Orientation Score), sufficient face resolution for detail (Resolution Score), and the consistent visibility of all facial parts (Completeness Score). A candidate must meet all five criteria, with further formulation details in Appendix~\ref{app:video_filtering_details}.

\subsubsection{Human Validation}

\paragraph{Protocol.} To confirm our automated filters serve as a reliable proxy for human quality judgments, we conduct a human evaluation study. For each of our seven filter criteria, we sample 100 borderline clips: 50 that marginally pass the filter and 50 that marginally fail. This focus on the decision boundary provides a stringent test of our thresholds. Two trained annotators, blind to the filter's decision, assign a binary label (e.g., ``Acceptable''/``Unacceptable'') to each clip based on a detailed rubric defining each quality attribute, more details can be found in Appendix~\ref{app:appendix_annotators}.

\paragraph{Results.} The evaluation confirms the reliability of our pipeline. First, we measure high inter-annotator agreement (IAA), achieving an average Cohen's Kappa ($\kappa$) of \textbf{0.79} across all criteria (Table~\ref{tab:kappa_scores_horizontal}). This indicates our quality standards are well-defined and consistently interpreted. Second, our automated filters demonstrate strong performance against the human-annotated ground truth, reaching an average accuracy of \textbf{95.1\%} and an F1-score of \textbf{95.3\%}. This result validates our automated pipeline as a reliable proxy for human quality assessment. Detailed per-criterion metrics are in the Appendix~\ref{subsec:appendix_table}. 

\begin{table}[t!]
\centering
\scriptsize 

\begin{tabular}{@{} l lcccccccr @{}} 
\toprule
Stage & CoT. & Dov. & Comp. & Move. & Orient. & Res. & Rot. & \textbf{Avg.} \\
\midrule
$\kappa$ & 0.74 & 0.90 & 0.80 & 0.66 & 0.80 & 0.72 & 0.90 & \textbf{0.79} \\
\bottomrule
\end{tabular}
\caption{Cohen's Kappa, $\kappa$ for quality filtering stages. Abbreviations are: CoTracker (CoT.), Dover (Dov.), Head-Completeness (Comp.), Head-Movement (Move.), Head-Orientation (Orient.), Head-Resolution (Res.), and Head-Rotation (Rot.).}
\label{tab:kappa_scores_horizontal}
\end{table}

\subsection{Quantitative Analysis}
\label{sec:dataset_analysis}



The final \texttt{TalkVid} dataset contains 1,244 hours of video from 7,729 unique speakers. Figure~\ref{fig:sankey_plot_for_data_filter} provides a holistic overview of the data attrition throughout our pipeline, while Figure~\ref{fig:data_distribution} details the statistical properties of the final dataset.\footnote{Data characteristics labeled as "Unknown" are not taken into account.}

\subsubsection{Composition}
Clips average 17.93s in duration. The dataset is compositionally diverse, led by ``Personal Experience'' (474.0h) and ``Vlogger/Creator'' (357.6h) content. It spans over 15 languages, with English (867.1h) and Chinese (248.9h) being the most prominent. Speaker demographics are varied across age, gender, and ethnicity, with the 31-45 year group being the largest (814.8h).

\subsubsection{Technical Quality}
Our filtering ensures high technical quality. A mean DOVER score of 8.55 confirms strong visual fidelity. The mean CoTracker ratio of 0.92 validates our selection for natural motion, successfully culling both overly static and erratic shots. Head detail scores (Movement, Rotation, Orientation) are sharply skewed towards their maxima, indicating stable, consistently trackable faces. This technical profile makes the dataset highly suitable for generative tasks.

\subsubsection{TalkVid-Core}
We introduce \textbf{TalkVid-Core}, a high-purity and diverse subset comprising 160 hours of content. This subset is derived by applying a stringent set of thresholds to quality metrics. Importantly, the data is uniformly sampled across ethnicity, gender, and age categories to ensure a balanced representation. Following the selection of this high-quality video set, we generate annotations for each clip using \textbf{Gemini 1.5 Pro}\footnote{gemini-1.5-pro-002}. Further details and qualitative examples of these annotations are provided in Appendix~\ref{app:annotation_visualization}.

\subsection{Qualitative Analysis}
\label{subsec:qualitative_analysis}

\begin{table*}[t]
\centering
\setlength{\tabcolsep}{1mm}  
\small  
\begin{tabular}{c|l|cccc|cccc|cccc}
\toprule
& \multirow{2}{*}{\shortstack[c]{\textbf{Training}\\\textbf{Dataset}}} & \multicolumn{4}{c|}{\textbf{English}} & \multicolumn{4}{c|}{\textbf{Chinese}} & \multicolumn{4}{c}{\textbf{Polish}} \\
\cmidrule(lr){3-6} \cmidrule(lr){7-10} \cmidrule(lr){11-14}
 & & \textbf{FID$\downarrow$} & \textbf{FVD$\downarrow$} & \textbf{Sync-C$\uparrow$} & \textbf{Sync-D$\downarrow$}
 & \textbf{FID$\downarrow$} & \textbf{FVD$\downarrow$} & \textbf{Sync-C$\uparrow$} & \textbf{Sync-D$\downarrow$} 
 & \textbf{FID$\downarrow$} & \textbf{FVD$\downarrow$} & \textbf{Sync-C$\uparrow$} & \textbf{Sync-D$\downarrow$} \\
\midrule
\multirow{3}{*}{\textbf{Language}} & HDTF & 60.817 & 443.202
& 4.000 & 10.403 & 48.561 & 415.564 & 3.285 & 10.058 & 39.231 & 321.261 & 2.654 & 10.368 \\
& Hallo3 & 59.842 & 387.507 & \textbf{4.753} & \textbf{9.536} & 52.514 & 342.062 & 4.005 & \textbf{9.450} & \textbf{38.458} & 343.553 & 3.424 & 9.690 \\
\rowcolor{gray!20} \cellcolor{white} & TalkVid & \textbf{59.562} & \textbf{357.603} & 4.567 & 9.867 & \textbf{47.509} & \textbf{306.131} & \textbf{4.041} & 9.521 & 39.271 & \textbf{288.178} & \textbf{3.695} & \textbf{9.663} \\
\midrule
 & & \multicolumn{4}{c|}{\textbf{White}} & \multicolumn{4}{c|}{\textbf{African}} & \multicolumn{4}{c}{\textbf{Asian}} \\
\midrule
\multirow{3}{*}{\textbf{Ethnicity}} & HDTF & 46.589 & 305.284 & 3.587 & 9.997 & 50.807 & 376.161 & 3.746 & 10.203
 & 53.163 & \textbf{302.214} & 3.453 & 10.198 \\
& Hallo3 & 40.927 & \textbf{267.492} & \textbf{4.218} & \textbf{9.292} & 48.218
& 350.025 & 4.296 & \textbf{9.636} & 52.492 & 303.478 & \textbf{4.076} & \textbf{9.517} \\
\rowcolor{gray!20} \cellcolor{white} & TalkVid & \textbf{40.740} & 274.226 & 4.176 & 9.587 & \textbf{44.373} & \textbf{326.840} & \textbf{4.352} & 9.724 & \textbf{48.511} & 303.997 & 4.056 & 9.744 \\
\midrule
 & & \multicolumn{4}{c|}{\textbf{Male}} & \multicolumn{4}{c|}{\textbf{Female}} & \multicolumn{4}{c}{\textbf{}} \\
\midrule
\multirow{3}{*}{\textbf{Gender}} & HDTF & 46.525 & 306.947 & 3.540 & 9.965 & 46.173 & 297.840 & 3.489 & 10.223  &  &  &  & \\
& Hallo3 & 41.549 & 299.984 & 3.935 & \textbf{9.496} & 42.659 & 258.583 & 4.034 & \textbf{9.704} & & & & \\
\rowcolor{gray!20} \cellcolor{white} & TalkVid & \textbf{39.398} & \textbf{294.709} & \textbf{3.984} & 9.639 & \textbf{41.967} & \textbf{241.920} & \textbf{4.051} & 9.788 \\
\midrule
 & & \multicolumn{4}{c|}{\textbf{19-30}} & \multicolumn{4}{c|}{\textbf{31-45}} & \multicolumn{4}{c}{\textbf{60+}} \\
\midrule
\multirow{3}{*}{\textbf{Age}} & HDTF & 45.591 & 283.927 & 3.679 & 10.078 & 58.843 & 295.236 & 3.592 & 10.448 & 53.192 & 350.580 & 3.630 & 10.018 \\
& Hallo3 & 41.501 & 272.912 & 4.214 & \textbf{9.565} & 44.493 & 253.756 & 4.380 & \textbf{9.581} & 53.854 & 332.383 & 3.748 & \textbf{9.741} \\
\rowcolor{gray!20} \cellcolor{white} & TalkVid & \textbf{37.879} & \textbf{253.698} & \textbf{4.329} & 9.605 & \textbf{43.702} & \textbf{222.202} & \textbf{4.537} & 9.626 & \textbf{51.141} & \textbf{321.556} & \textbf{3.942} & 9.804 \\
\bottomrule
\end{tabular}
\caption{Comparison with other baseline training datasets, including HDTF~\cite{zhang2021flow} and Hallo3~\cite{cui2024hallo3} on \texttt{TalkVid-bench} across four dimensions, showing subgroup-level performance.}
\label{tab:compare_dataset_talkvid_bench_fine}
\end{table*}

\begin{figure}[t]           
    \centering
    \includegraphics[width=\columnwidth]{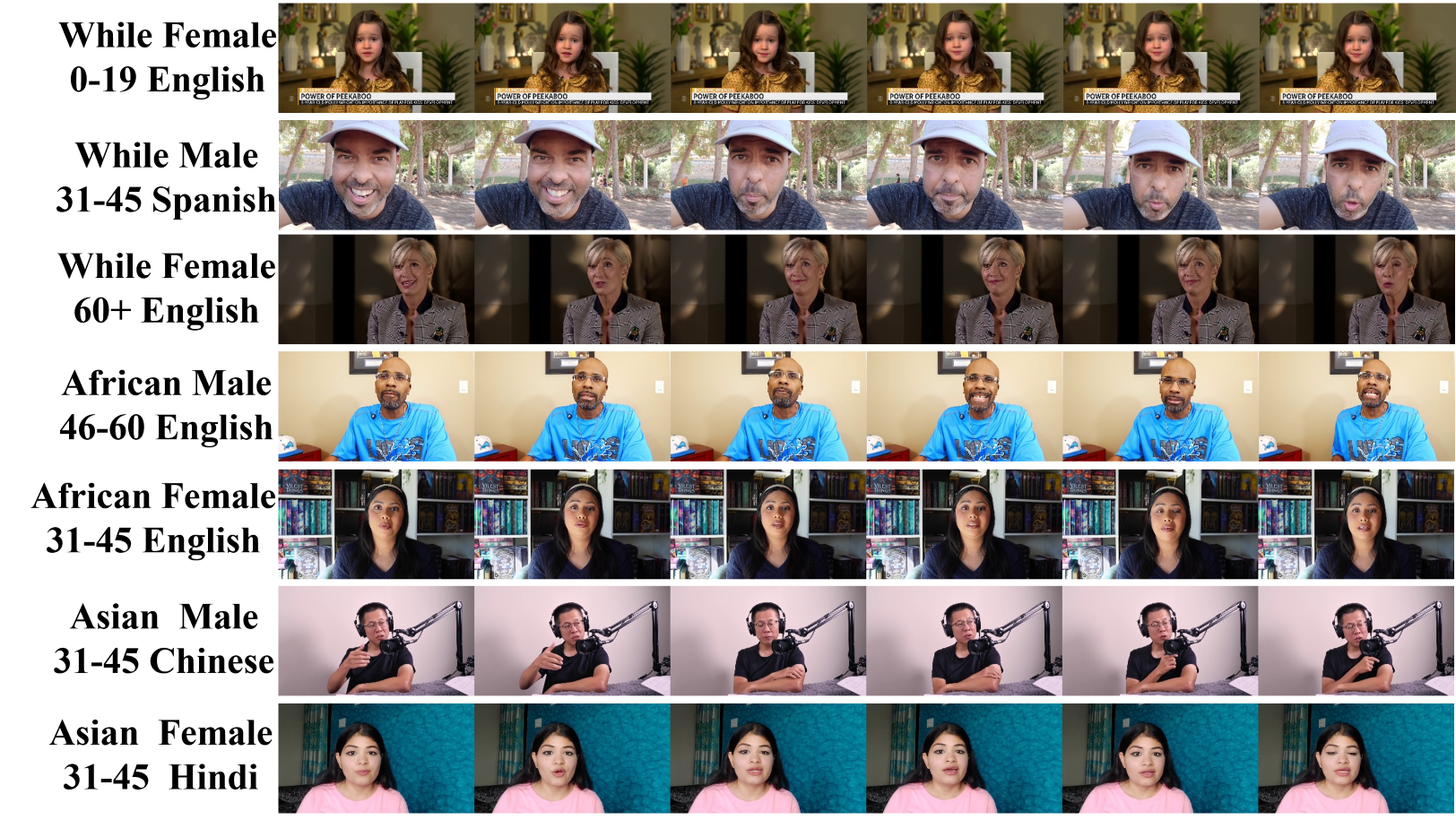}
    \caption{Qualitative examples from \textbf{TalkVid}. The sequences illustrate demographic diversity and key technical challenges for synthesis: varied lighting, complex backgrounds, and occlusions.}
    \label{fig:case_study_grids}
\end{figure}

Figure~\ref{fig:case_study_grids} illustrates the dataset's breadth. The examples confirm the demographic diversity (age, gender, ethnicity) and show difficult capture conditions. These include in-the-wild lighting (Row 1), complex indoor backgrounds (Row 4), and accessories such as headwear and glasses (Rows 1, 5). The sequences also contain significant challenges for synthesis models, including occlusions from hands and microphones (Row 5) and a full range of motion dynamics, from large-amplitude expressions (Row 1) to subtle conversational movements (Row 2). The inclusion of diverse subjects under these conditions makes \texttt{TalkVid} a valuable resource for training and evaluating generative models.

\begin{table*}[t]
\centering
\small  
\begin{tabular}{lcccccccc}
\toprule
& \multicolumn{4}{c}{\textbf{Language}} & \multicolumn{4}{c}{\textbf{Ethnicity}}\\
\cmidrule(lr){2-5} \cmidrule(lr){6-9}
Training Dataset
  & FID$\downarrow$ & FVD$\downarrow$ & Sync-C$\uparrow$ & Sync-D$\downarrow$ & FID$\downarrow$ & FVD$\downarrow$ & Sync-C$\uparrow$ & Sync-D$\downarrow$ \\
\midrule
HDTF &31.385 &205.990 &3.109 &10.567 &36.381 &214.488 &3.605 &10.135 \\ 
Hallo3 &29.721 & 184.465 &\textbf{3.849} &\textbf{9.887} &34.650 &194.847 &4.204 &\textbf{9.488} \\
\rowcolor{gray!20} TalkVid & \textbf{28.686} &\textbf{178.396} &3.842 &10.064 &\textbf{32.588} &\textbf{187.368} &\textbf{4.205} &9.686 \\
\midrule
& \multicolumn{4}{c}{\textbf{Gender}} & \multicolumn{4}{c}{\textbf{Age}}\\
\cmidrule(lr){2-5} \cmidrule(lr){6-9}
Training Dataset
  & FID$\downarrow$ & FVD$\downarrow$ & Sync-C$\uparrow$ & Sync-D$\downarrow$ & FID$\downarrow$ & FVD$\downarrow$ & Sync-C$\uparrow$ & Sync-D$\downarrow$ \\
\midrule
HDTF &38.950 &225.611 &3.513 &10.100 &34.257 &166.276 &3.542 &10.168\\ 
Hallo3 &35.260 &208.815 &3.987 &\textbf{9.605} &32.934 &158.799 &4.052 &\textbf{9.596} \\
\rowcolor{gray!20} TalkVid &\textbf{34.347} &\textbf{199.003} &\textbf{4.019} &9.717 &\textbf{30.790} &\textbf{151.580} &\textbf{4.112} &9.731 \\
\bottomrule
\end{tabular}
\caption{Comparison with other baseline training datasets, including HDTF~\cite{zhang2021flow} and Hallo3~\cite{cui2024hallo3} on \texttt{TalkVid-bench} across four dimensions in general.}
\label{tab:compare_dataset_talkvid_bench}
\end{table*}
\section{Experiments}
\label{sec:experiments}

We conduct experiments to validate the benefits of \texttt{TalkVid} as a training corpus and to demonstrate the utility of \texttt{TalkVid-Bench} for revealing model-specific biases. We compare a SOTA model trained on \texttt{TalkVid} against the same model trained on prior datasets.

\subsection{Experimental Setup}

\subsubsection{Model and Baselines} We train the open-source V-Express~\cite{wang2024v} model, a SOTA diffusion-based architecture for talking-head synthesis. We evaluate its performance when trained under three distinct dataset conditions: 1) \textbf{HDTF}~\cite{zhang2021flow}: A high-resolution talking-head dataset. 2) \textbf{Hallo3}~\cite{cui2024hallo3}: A curated dataset with clean motion conditions. 3) \textbf{TalkVid-Core} (Ours): A 160-hour subset of our proposed \texttt{TalkVid} dataset.

\subsubsection{Implementation Details} For all conditions, we adhere strictly to the original three-stage training protocol of V-Express (40k, 75k, and 50k steps) and its hyperparameters. We use the AdamW optimizer~\citep{loshchilov2018decoupled} with a learning rate of 1e-6 and global batch sizes of 8, 4, and 2 for each stage, respectively. Input video frames are preprocessed by cropping facial regions and resizing to \textbf{512×512}. Training for each condition requires 3 days on 4 NVIDIA A100 GPUs.

\subsubsection{Evaluation datasets}
\label{sec:benchmark}

Evaluation is conducted on three test sets. First, we use a 100-clip subset of the HDTF test set. Second, we use a 167-clip subset from the Hallo3 test set to evaluate performance on cleaner motion and larger pose variations.
Third, we introduce \texttt{TalkVid-Bench}, our primary benchmark of 500 five-second clips designed for robust and fair evaluation. Drawn from the \texttt{TalkVid} corpus but held-out from training, the benchmark is stratified and balanced across four dimensions: language, ethnicity, gender, and age. Whereas HDTF and Hallo3 only support aggregate scores, \texttt{TalkVid-Bench} enables granular analysis of model performance across subgroups, making it the definitive tool for the cross-domain and fairness experiments in this paper. Subgroup distributions are detailed in Appendix~\ref{app:benchmark_stats}.

\subsubsection{Evaluation metrics}
We employ a set of metrics to evaluate performance:
\textbf{Visual Quality.} We report Frechet Inception Distance (\textbf{FID})~\cite{heusel2017gans} for per-frame realism and Frechet Video Distance (\textbf{FVD})~\cite{unterthiner2018towards} for temporal coherence and video-level fidelity.
\textbf{Audio-Visual Synchronization.} Following SyncNet~\cite{chung2016syncnet}, we measure audio-lip sync confidence (\textbf{Sync-C}) and the distance between audio-visual embeddings (\textbf{Sync-D}).

\subsection{Quantitative Results
}

\subsubsection{Fine-grained Results on TalkVid-Bench}
The fine-grained evaluation on \texttt{TalkVid-Bench}, shown in Table~\ref{tab:compare_dataset_talkvid_bench_fine}, confirms that \texttt{TalkVid} produces models with superior generalization and reduced bias.

\paragraph{Cross-lingual Generalization} While baselines perform well on English, our model achieves the best visual quality (FID/FVD) across English, Chinese, and Polish, significantly outperforming on non-English languages. This result validates that \texttt{TalkVid}'s linguistic breadth directly remedies the brittleness of models trained on narrower data.

\paragraph{Mitigating Ethnic Bias} The Hallo3-trained model is competitive for White speakers but falters for African speakers, where our model is clearly superior. This performance delta, revealed by \texttt{TalkVid-Bench}, shows that TalkVid's inclusive data fosters more equitable models.

\paragraph{Robustness Across Gender and Age} Our model achieves the most consistent high performance for both male and female subjects and shows marked improvements for challenging age groups, particularly for speakers aged 60+. This underscores the value of \texttt{TalkVid}'s comprehensive demographic coverage.


\subsubsection{Overall Results on TalkVid-Bench}
Table~\ref{tab:compare_dataset_talkvid_bench} shows the overall performance for each dimension of \texttt{TalkVid-Bench}. Three trends stand out.

\paragraph{Visual fidelity leads across the board} \texttt{TalkVid} consistently records the lowest FID/FVD in all four dimensions, confirming that its higher-quality training data translates into universally sharper and more stable videos.

\paragraph{Synchronisation remains competitive} Hallo3 consistently achieves the lowest Sync-D, but the gaps are small. \texttt{TalkVid} matches or surpasses Hallo3 on Sync-C in three of the four dimensions and is only marginally lower on Language, showing that visual improvements do not come at a meaningful cost to lip-sync quality.

\paragraph{Balanced demographic performance} From language and ethnicity to gender and age, \texttt{TalkVid} delivers the most uniform improvements, reinforcing the conclusion that a diverse training set yields a model that generalises well without introducing new biases.


\begin{table}[t]
\centering
\setlength{\tabcolsep}{1mm}  
\small  
\begin{tabular}{lcccc}
\toprule
& \multicolumn{4}{c}{HDTF test set} \\
\cmidrule(lr){2-5}
Training Dataset
  & FID$\downarrow$ & FVD$\downarrow$ & Sync-C$\uparrow$ & Sync-D$\downarrow$ \\
\midrule
HDTF &\textbf{19.963} &188.657 &3.523 &10.254 \\ 
Hallo3 &24.484 & 183.172 & 3.615 &\textbf{9.966} \\
\rowcolor{gray!20} 
TalkVid & 21.772 &\textbf{175.122} &\textbf{3.707} & 9.970 \\
\midrule
& \multicolumn{4}{c}{Hallo3 test set} \\
\cmidrule(lr){2-5}
Training Dataset 
  & FID$\downarrow$ & FVD$\downarrow$ & Sync-C$\uparrow$ & Sync-D$\downarrow$ \\
\midrule
HDTF &\textbf{16.690} &114.891 &4.464 &9.901\\ 
Hallo3 &19.745 & 106.009 &\textbf{4.941} &\textbf{9.404}\\
\rowcolor{gray!20} 
TalkVid  & 18.367 &\textbf{101.819} & 4.921 & 9.530 \\
\bottomrule
\end{tabular}
\caption{Comparison with other baseline training datasets on HDTF and Hallo3 test sets.}
\label{tab:compare_dataset_hdtf_hallo3}
\end{table}


\subsubsection{Comparison on Standard Benchmarks}

We further assess generalization on the canonical HDTF and Hallo3 test sets (Table~\ref{tab:compare_dataset_hdtf_hallo3}). The TalkVid-trained model demonstrates superior cross-domain robustness, achieving the best temporal coherence (FVD) on \emph{both} benchmarks and strong lip-sync (Sync-C) scores. In contrast, models trained on HDTF and Hallo3 exhibit significant performance degradation when evaluated out-of-domain, highlighting their tendency to overfit. Notably, while the HDTF model secures the best FID on its own test set, it does so at the cost of the worst FVD. Our model makes a more effective trade-off, indicating that TalkVid's diversity promotes a more balanced and generalizable synthesis that avoids overfitting to domain-specific visual artifacts.

\begin{figure}[t]
\centering
\includegraphics[width=\linewidth]{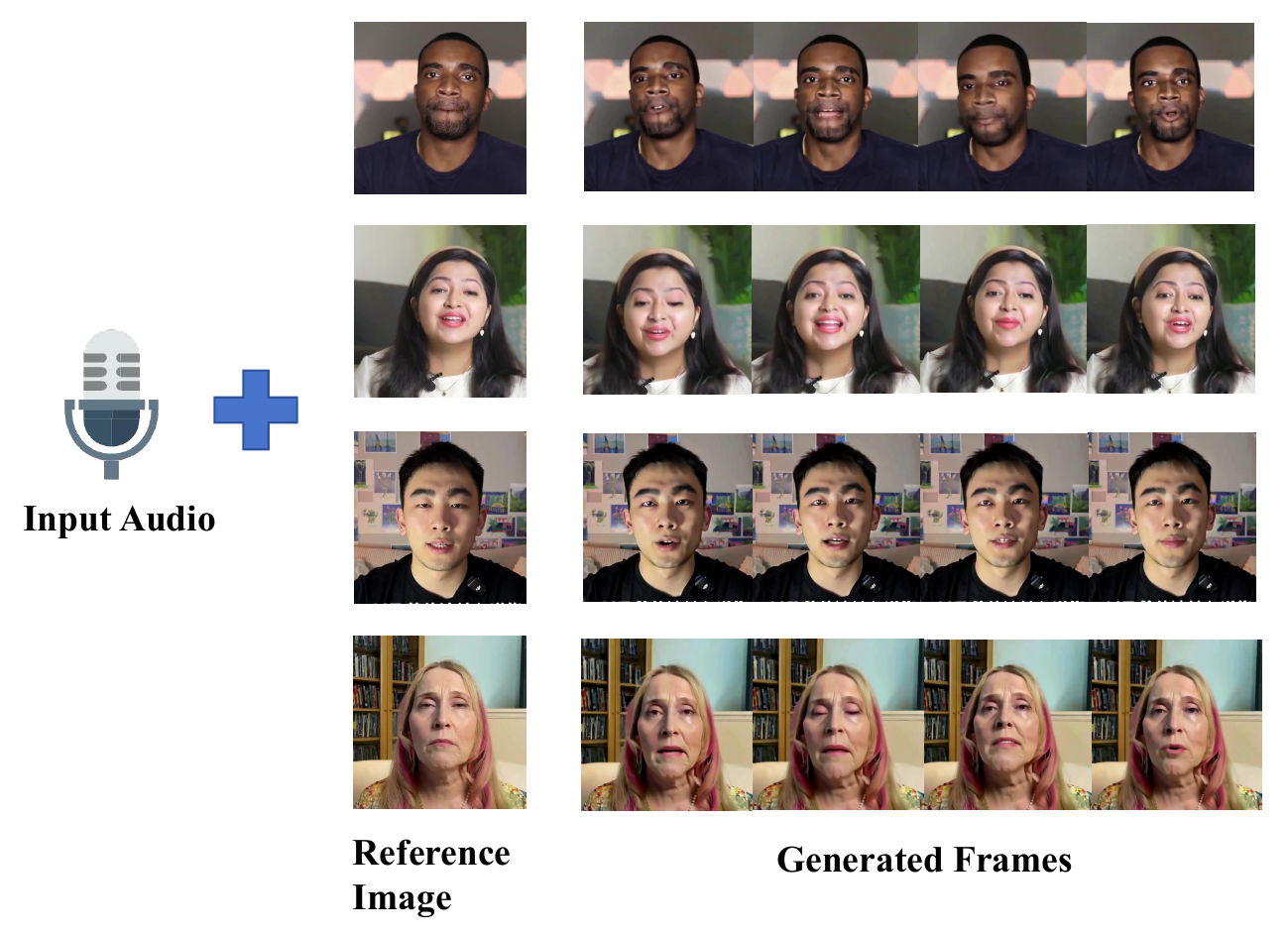}
\caption{Qualitative examples from our TalkVid-trained model, evaluated on diverse samples from \texttt{TalkVid-Bench} spanning language, ethnicity, gender, and age.}
\label{fig:demo_1}
\end{figure}

\begin{figure}[t]
\centering
\includegraphics[width=\linewidth]{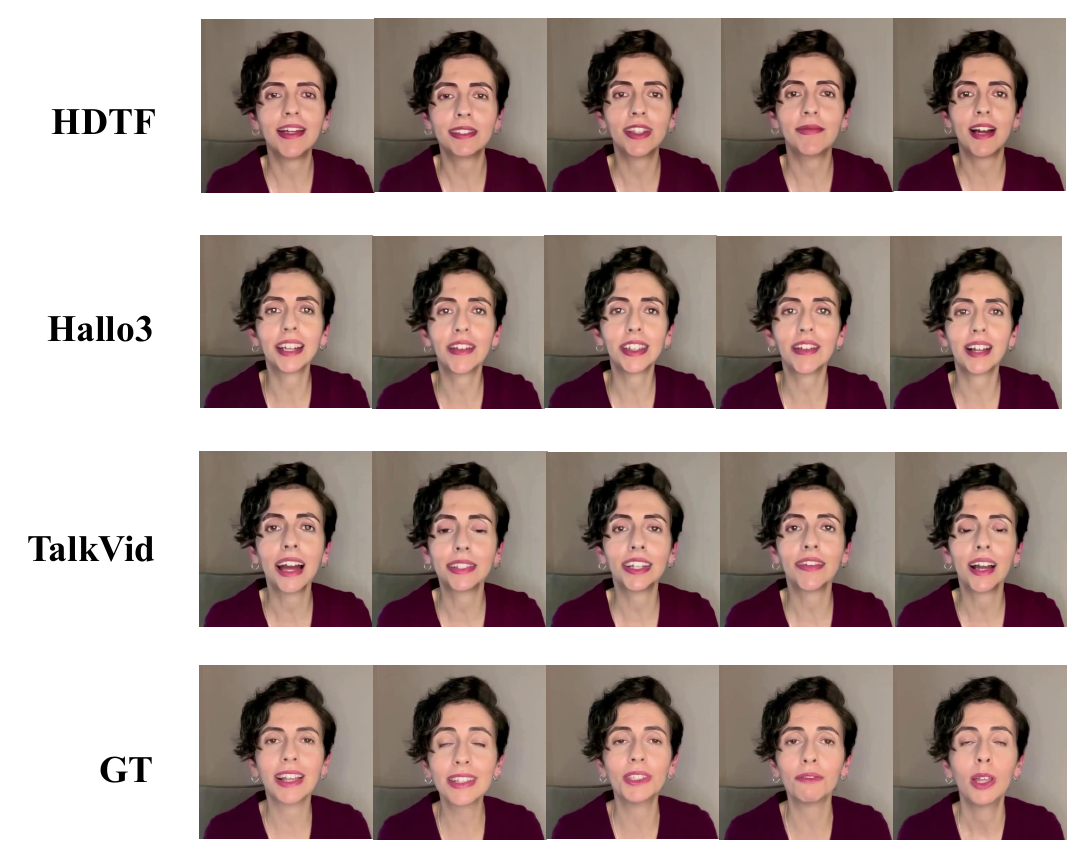}
\caption{Qualitative comparison on an unseen clip from \texttt{TalkVid-Bench}. From top: V-Express fine-tuned on HDTF, Hallo3, \emph{TalkVid-Core} (ours), and Ground-Truth (GT).}
\label{fig:demo_2}
\end{figure}

Training on \texttt{TalkVid} yields a model that is robust across evaluation domains and metrics. The quantitative results, particularly the stratified analysis on \texttt{TalkVid-Bench}, prove that \texttt{TalkVid} provides a superior foundation for developing talking-head models that are both high-fidelity and equitable.

\subsection{Qualitative Results}

\subsubsection{Diversity coverage and naturalness}
Figure~\ref{fig:demo_1} showcases the performance of our model on diverse identities from \texttt{TalkVid-Bench}. The model accurately preserves identity and background across speakers. Crucially, it synthesizes natural, non-verbal behaviors often absent in prior work: subtle head motion (row 2) and realistic eye blinks (row 4) are generated in sync with speech. This confirms that the model generalizes well beyond near-frontal, static poses.

\subsubsection{Comparison with baseline datasets}
The frame-by-frame comparison in Figure~\ref{fig:demo_2} reveals clear deficiencies in the baseline models. Both the HDTF and Hallo3-trained models produce static expressions with muted lip motion, failing to match the audio or replicate natural behaviors like eye blinks. In contrast, our model reproduces the ground-truth's dynamic expression, including correctly timed eye blinks and larger, more articulate lip shapes. This qualitative evidence corroborates our quantitative findings, confirming that \texttt{TalkVid}'s rich motion diversity leads to more lifelike and accurate synthesis.
\section{Ethical Considerations}
\label{sec:ethics}
While generative models pose significant risks of misuse, we contend that an equally critical ethical failure is the status quo: creating biased technology from non-diverse data that systematically fails for underrepresented groups. Our work directly confronts this harm. \texttt{TalkVid} provides the demographically rich data to train fairer models, while \texttt{TalkVid-Bench} offers a standardized framework to audit and mitigate such algorithmic bias. To ensure responsible dissemination, we will distribute the dataset as source URLs and timestamps to verified researchers under a strict license. This protocol respects creator copyright and explicitly prohibits all malicious applications, including defamation and non-consensual content generation, thereby balancing research accessibility with accountability.

\section{Conclusion}
This paper addressed the critical brittleness of SOTA talking head models, a direct consequence of inadequate training data. We introduced \texttt{TalkVid}, a large-scale, diverse, and technically pristine dataset curated through a rigorous, human-validated pipeline. To enable fair assessment, we also presented \texttt{TalkVid-Bench}, a stratified benchmark that uncovers biases invisible to standard metrics. Our experiments show that training on \texttt{TalkVid} produces more robust and equitable models. By releasing this ecosystem, we hope it will spur further research into auditing and mitigating bias in generative video models.

\clearpage

{
    \small
    \bibliographystyle{ieeenat_fullname}
    \bibliography{main}
}

\clearpage
\appendix

\section{Video Collection Criteria}
\label{app:video_collection_criteria}
To ensure consistency and quality in the collected data, we establish standardized guidelines across three key dimensions: recording environment, speaker behavior, and identity diversity.

\textbf{Recording Environment Requirements.} All videos are recorded indoors to avoid uncontrollable outdoor factors such as variable lighting or wind noise. Lighting conditions must be stable and evenly distributed, with strong side or backlighting strictly avoided. The background should be visually clean and preferably monochromatic to minimize distractions. Recording devices are required to support a minimum resolution of 1080p and a frame rate of at least 25\, fps, and should be mounted stably—preferably on a tripod—to avoid camera shake or motion blur. Audio must be captured clearly, and free from background noise or interference (e.g., music, environmental sounds). The recorded audio should contain only a single speaker, with no overlapping dialogue or ambient speech.

\textbf{Speaker Movement Requirements.} During recording, speakers are instructed to face the camera directly, maintaining a natural and relaxed facial expression. Excessive head motion, exaggerated gestures, or sudden movements are discouraged to preserve alignment quality. The speaker’s face should remain unobstructed throughout the recording—masks, microphones, or large reflective glasses are not allowed, while standard eyeglasses are acceptable. The entire facial region, including the chin and forehead, must stay within the camera frame, with the face occupying approximately 30–40\% of the frame area. Appropriate headroom and a consistent shooting distance (recommended 0.5–1 meter) should be maintained. Speech content must be delivered in clear, accent-neutral English at a moderate pace, with well-pronounced articulation.

\textbf{Speaker Diversity Requirements.} To promote fairness and generalizability in downstream applications, the dataset is curated to include a diverse range of speakers. We ensure balanced representation across genders, age groups, ethnic backgrounds, and speaking styles. Collected samples vary in facial expressions, emotional tone, speaking speed, and prosody. Each video features unique spoken content between 10 to 30 seconds in duration, avoiding repetition or overly scripted delivery. For quality assurance, collectors verify facial visibility and audiovisual synchronization and record basic demographic metadata (e.g., gender, age group, race).

\section{Video Filtering Details}
\label{app:video_filtering_details}
{\subsection{Motion Filtering Details}}
We adopt the point-tracking stability ratio $\rho$ provided by CoTracker~\cite{karaev2024cotracker3} as a proxy for natural facial motion. 
For each 16-frame clip, CoTracker initializes $K=256$ trajectories on a uniform grid within the central $256\times256$ crop. 
A trajectory is deemed \emph{valid} over the entire clip if: 1) its confidence remains $\geq 0.5$ in every frame, 2) the per-frame displacement never exceeds 20 pixels (approx.~8\% of the image diagonal), 3)it stays inside the frame boundaries.
The stability ratio is then defined as:
\begin{equation}
    \rho = \frac{\#\text{valid trajectories}}{K}\in[0,1].
\end{equation}
A high $\rho$ indicates consistent tracking but not necessarily large motion; conversely, a low $\rho$ usually corresponds to motion blur or tracking failure.
We retain clips whose stability ratio satisfies
\begin{equation}
    0.85 \leq \rho \leq 0.999.
\end{equation}

\textbf{Lower bound ($\rho\geq0.85$).} 
Clips with $\rho<0.85$ exhibit $>10\%$ trajectory loss, typically caused by severe motion blur or compression artifacts that would corrupt subsequent 3D landmark estimation.

\textbf{Upper bound ($\rho\leq0.999$).} 
Clips with $\rho>0.999$ contain almost no detectable micro-motion (empirical mean displacement $<0.3$ px), resulting in unnaturally static ``frozen'' faces that degrade the perceived liveliness of generated talking heads. All videos are resampled to 25 fps and down-scaled so that the shorter side is 512 px. We extract 16-frame clips with a sliding window stride of 8 frames.

\subsection{Head-detail Filtering Details}
\label{app:head_filter}
To evaluate the facial quality in video clips, we define a scoring system based on five dimensions. All scores range from 0 to 100, with the exception of the \textbf{Resolution Score} which ranges from 0 to 3000. Higher values indicate better quality for all scores. A clip is retained only if all scores meet their respective thresholds.

\paragraph{Movement Score.}
This metric measures the temporal stability of facial keypoints. We compute the average displacement of keypoints between adjacent frames, normalized by the smaller dimension of the image (height or width). The score is defined as $100 - 100 \times \text{avg\_movement}$. Lower displacement leads to higher scores. \textbf{Threshold:} average $\geq 80$, minimum $\geq 60$.

\paragraph{Orientation Score.}
This score reflects how frontal the face is across frames. For each frame, we compute pitch, yaw, and roll scores as $|\theta| / 180 \times 100$, and derive the final score as $100 - \sqrt{\text{pitch}^2 + \text{yaw}^2 + \text{roll}^2}$. A higher score indicates better alignment with the camera. \textbf{Threshold:} average $\geq 70$, minimum $\geq 30$.

\paragraph{Completeness Score.}
We assess whether key facial regions are fully visible within the frame. The score combines three regions: eyes (weight 0.3), nose (0.4), and mouth (0.3). For each region, the presence of all keypoints within image bounds contributes 1; otherwise, 0. The weighted sum gives the final score. Occlusions are tolerated as long as keypoints are within the visible area. \textbf{Threshold:} average $= 100$, minimum $= 100$.

\paragraph{Resolution Score.}
This score quantifies how large the face appears in the frame. It is calculated as $30 \times (\text{face\_area} / \text{image\_area}) \times 100$. Larger face regions yield higher scores, which may exceed 100. \textbf{Threshold:} average $\geq 50$, minimum $\geq 40$.

\paragraph{Rotation Score.}
This metric evaluates the smoothness of head motion. It is defined as $100 - \text{avg\_rotation\_amplitude}$, where the amplitude is computed by the 3D orientation change: $\sqrt{\Delta \text{pitch}^2 + \Delta \text{yaw}^2 + \Delta \text{roll}^2}$ between adjacent frames. Smaller variations indicate higher stability. \textbf{Threshold:} average $\geq 70$, minimum $\geq 60$.

\noindent All metrics must satisfy their corresponding thresholds for a clip to pass the quality screening.

\section{Video Annotation}

\label{sec:appendix_eval_details}

This appendix provides supplementary details for the human verification study discussed in the main paper.

\subsection{Detailed Performance Metrics}
\label{subsec:appendix_table}

A detailed breakdown of the evaluation results for each of the seven automated filtering stages is summarized in Table~\ref{tab:human_eval_details}. For each category, the table presents the Inter-Annotator Agreement (IAA) rate, which measures annotator consistency, alongside key classification metrics (Accuracy, Precision, Recall, and F1-score) for our automated filter when benchmarked against the Golden Standard. The consistently high scores reported in the table underscore the robustness and reliability of our data curation pipeline.

\begin{table*}[ht]
\centering
\begin{tabular}{l c c c c c}
\toprule
\textbf{Filtering Stage} & \textbf{IAA} & \textbf{Accuracy} & \textbf{Precision} & \textbf{Recall} & \textbf{F1-score} \\
\midrule
Cotracker           & 86.87\% & 87.21\% & 88.64\% & 86.67\% & 87.64\% \\
Dover               & 94.00\% & 96.81\% & 95.83\% & 97.87\% & 96.84\% \\
Head Completeness   & 90.00\% & 96.67\% & 93.48\% & 100.00\% & 96.63\% \\
Head Movement       & 83.00\% & 96.39\% & 92.50\% & 100.00\% & 96.10\% \\
Head Orientation    & 90.00\% & 94.44\% & 97.78\% & 91.67\% & 94.62\% \\
Head Resolution     & 86.00\% & 97.67\% & 94.59\% & 100.00\% & 97.22\% \\
Head Rotation       & 95.00\% & 96.84\% & 95.83\% & 97.87\% & 96.84\% \\
\midrule
\textbf{Average}    & \textbf{89.3\%} & \textbf{95.1\%} & \textbf{95.5\%} & \textbf{96.3\%} & \textbf{95.3\%} \\
\bottomrule
\end{tabular}
\caption{Detailed Human Evaluation Results for Each Filtering Stage. We report Inter-Annotator Agreement (IAA) and the classification performance (Accuracy, Precision, Recall, F1-score) of our automated filter against the Golden Standard.}
\label{tab:human_eval_details}
\end{table*}

\subsection{Annotator Background and Training}
\label{app:appendix_annotators}

The human evaluation was conducted by a team of five annotators with strong technical and scientific backgrounds. The team comprised two Ph.D. students in Computer Science, one Ph.D. student in Applied Mathematics, one undergraduate student in Computer Science, and one undergraduate student in Statistics. All evaluators possess substantial experience with rigorous scientific research methodologies.

To ensure consistency and objectivity, a strict evaluation protocol was enforced. For each of the seven filtering categories, a pair of these annotators was assigned to evaluate the 100 sample clips independently. Prior to the main annotation task, all participants underwent a dedicated calibration session. During this session, they were provided with detailed written guidelines and illustrative examples for each quality criterion (e.g., defining acceptable vs. unacceptable head movement). The goal of this phase was to establish a shared and consistent understanding of the task requirements, which directly contributed to the high inter-annotator agreement rates observed in our results. 

Crucially, the entire evaluation process was \textbf{blinded}; the annotators had no knowledge of the decisions made by the automated filters, ensuring that their judgments remained completely unbiased.

\subsection{Annotation Guidelines for Human Evaluators}
\label{subsec:appendix_guidelines}

To ensure all annotators applied the same criteria, we provided them with the following detailed instructions for each filtering category. These guidelines were also used during the calibration session to resolve ambiguities.

\subsubsection{General Rule}
A critical edge case applied to all categories: clips where a human face or head could not be reliably detected were generally to be labeled as \textbf{negative}, even if the clip otherwise met the quality criterion (e.g., high video quality or smooth motion).

\subsubsection{Filter-Specific Instructions}
\begin{description}
    \item[Cotracker Filter] The primary criterion is the spatio-temporal stability of the head. Clips were labeled as \textbf{negative} if they contained sudden, jerky movements resulting from large translations or rapid 3D rotations of the head.

    \item[Dover Filter] This filter assesses overall visual and technical quality. Clips were labeled as \textbf{negative} if they suffered from low resolution, significant compression artifacts, poor lighting, or motion blur.

    \item[Head Detail Filters] This is a composite evaluation. A clip must generally meet all five of the following sub-criteria to be labeled as \textbf{positive}:
    \begin{itemize}
        \item \textbf{Movement Stability:} The head must remain relatively stationary. Clips with large, abrupt translations or occlusions were rejected.
        \item \textbf{Frontal Orientation:} The subject's face must be predominantly front-facing. Clips containing significant head turns, downward gazes, or profiles were rejected.
        \item \textbf{Head Completeness:} The entire facial region must be clearly visible and unobstructed. Clips showing only partial features or where the face was occluded were rejected.
        \item \textbf{Facial Resolution:} The face must occupy a salient portion of the frame (heuristically, $>20\%$). Clips where the face was too small (e.g., $<10\%$ of the frame area) were rejected.
        \item \textbf{Rotational Stillness:} This criterion is exceptionally strict. The head must maintain a fixed orientation with minimal rotation. Even a single, noticeable head turn, nod, or shake within the clip was sufficient for it to be labeled as \textbf{negative}. Annotators were instructed to watch the entire clip before making a final judgment.
    \end{itemize}
\end{description}

\begin{figure*}[t]
    \centering
    \includegraphics[width=0.9\linewidth]{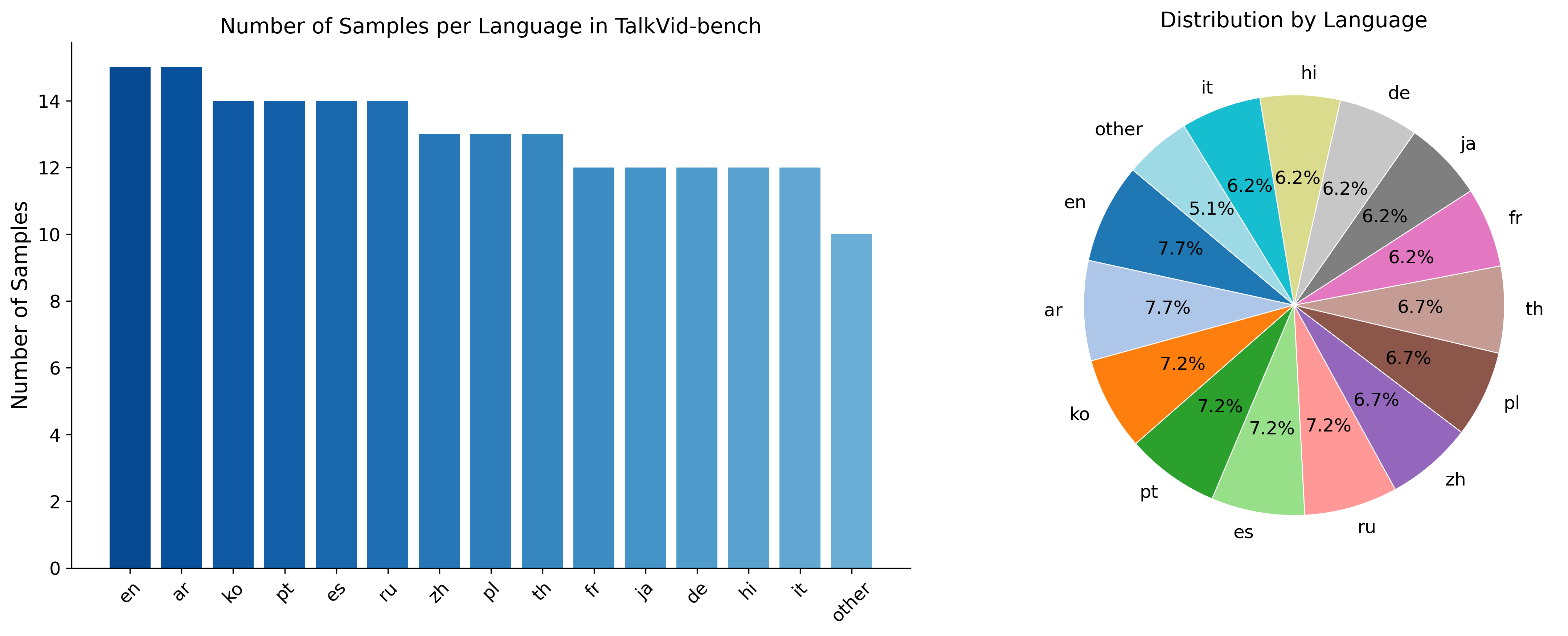}
    \caption{Distribution of \textit{TalkVid-bench} across the language dimension (15 languages, 195 samples). The left panel shows the number of samples per language, while the right panel shows the distribution of these samples by language. Language abbreviations: \textbf{ar} (Arabic), \textbf{pl} (Polish), \textbf{de} (German), \textbf{ru} (Russian), \textbf{fr} (French), \textbf{ko} (Korean), \textbf{pt} (Portuguese), \textbf{other} (other languages), \textbf{ja} (Japanese), \textbf{th} (Thai), \textbf{es} (Spanish), \textbf{it} (Italian), \textbf{hi} (Hindi), \textbf{en} (English), \textbf{zh} (Chinese).}
    \label{fig:benchmark_dist_lang} 
\end{figure*}

\begin{figure*}[t]
    \centering
    \includegraphics[width=0.9\linewidth]{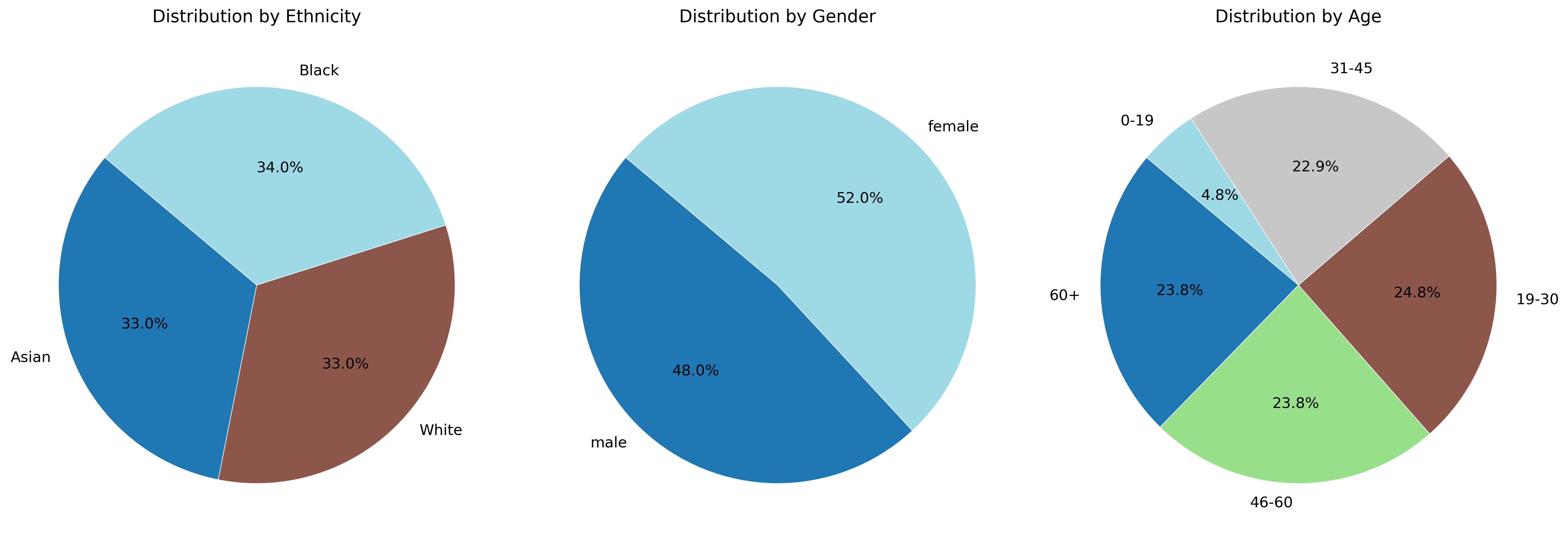}
    \caption{Distribution of \textit{TalkVid-bench} across three demographic dimensions.
These statistics illustrate the diversity of TalkVid-bench in terms of participant demographics,
providing a comprehensive benchmark for evaluating models under varied demographic conditions.}
    \label{fig:benchmark_dist_demo} 
\end{figure*}

\subsubsection{Visual Examples}
To further clarify the annotation criteria, this section provides visual examples. We first show an ideal \textbf{positive} case that passes all filters. Subsequently, we present seven \textbf{negative} cases, each illustrating a failure for a specific filtering criterion. Annotators were instructed to label the following types of clips as \texttt{negative}.

\subsection{Benchmark Design}
\label{app:benchmark_stats}
TalkVid-Bench comprises 500 carefully sampled and stratified video clips along four critical demographic and language dimensions: \textbf{age}, \textbf{gender}, \textbf{ethnicity}, and \textbf{language}. This stratified design enables granular analysis of model performance across diverse subgroups, mitigating biases hidden in traditional aggregate evaluations. Each dimension is divided into balanced categories:
\begin{itemize}
    \item \textbf{Age:} 0--19, 19--30, 31--45, 46--60, 60+, with a total of 105 samples.
    \item \textbf{Gender:} Male, Female, with a total of 100 samples.
    \item \textbf{Ethnicity:} Black, White, Asian, with a total of 100 samples.
    \item \textbf{Language:} English, Chinese, Arabic, Polish, German, Russian, French, Korean, Portuguese, Japanese, Thai, Spanish, Italian, Hindi, and Other languages, with a total of 195 samples.
\end{itemize}

\section{Computational Efficiency}
We quantify efficiency using the \emph{real-time factor} (RTF), defined as the ratio between the input-video duration and the wall-clock processing time. An RTF greater than~1 indicates faster-than-real-time operation. Our pipeline comprises following sequential stages:

\begin{enumerate}
    \item \textbf{Rough segmentation + subtitle filtering} (CPU-only, 96~cores) achieves an average RTF of $\mathbf{18.14}$.
    \item \textbf{Motion filtering (CoTracker)} (96-core CPU with $8\times$\,NVIDIA~A800 GPUs) reaches an average RTF of $\mathbf{64.21}$.
    \item \textbf{Quality filtering (DOVER)} (96-core CPU with $8\times$\,NVIDIA~A800 GPUs) reaches an average RTF of $\mathbf{87.36}$.
    \item \textbf{Head filtering} (96-core CPU with $8\times$\,NVIDIA~A800 GPUs) reaches an average RTF of $\mathbf{72.47}$.
\end{enumerate}

\newcolumntype{M}[1]{>{\centering\arraybackslash}m{#1}}

\begin{table*}[htbp]
\centering
\setlength{\tabcolsep}{6pt}      
\small
\begin{tabular}{@{}M{1.5cm}M{11.5cm}@{}}
\toprule
\multicolumn{2}{@{}>{\centering\arraybackslash}m{13cm}@{}}
{\textbf{Examples of Sample Quality Based on Filter Cotracker}}\\
\midrule
\textbf{Quality Level} & \textbf{Sample Example}\\
\midrule
Poor & \includegraphics[width=0.95\linewidth]{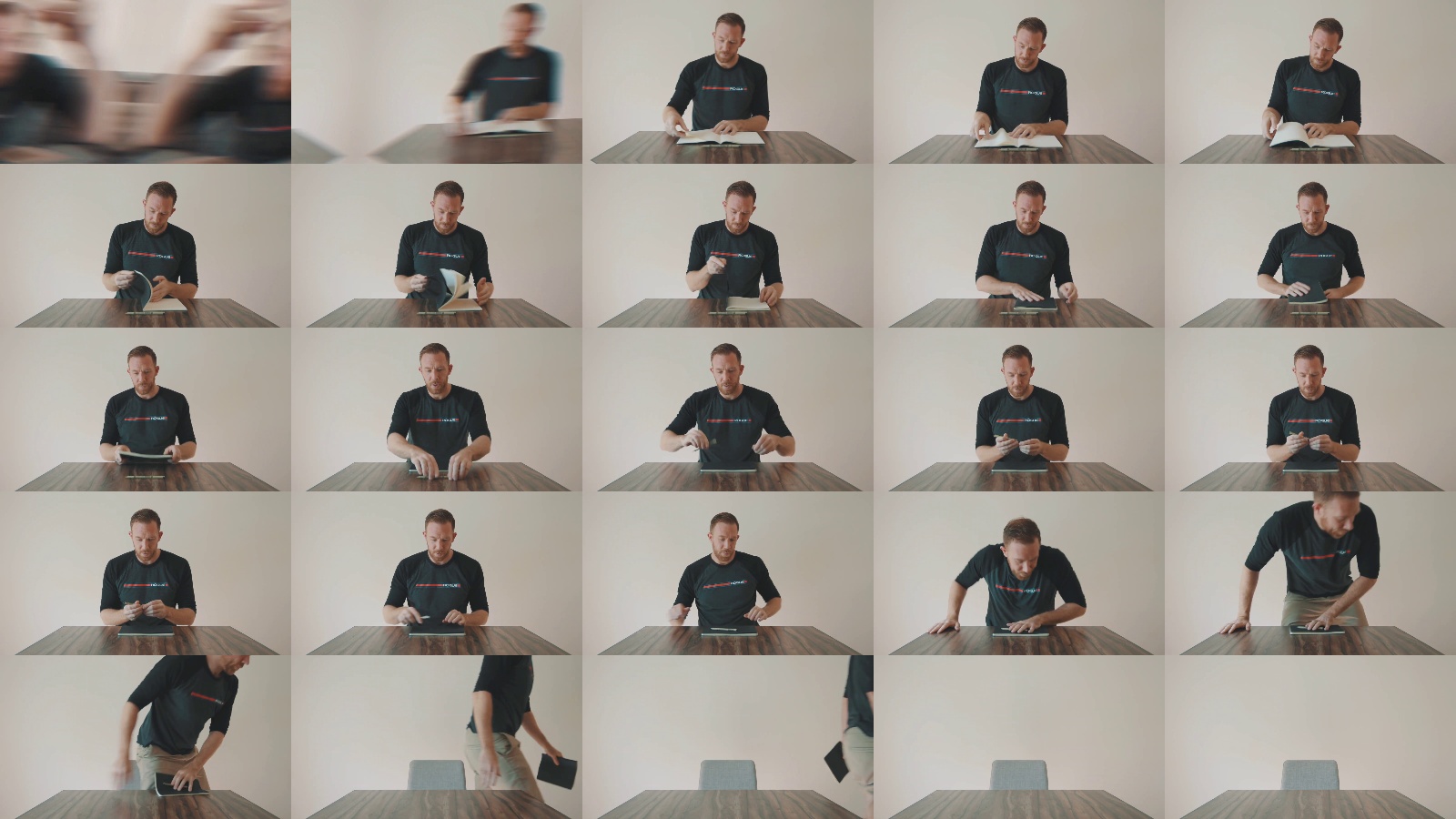}\\
\midrule
Fair & \includegraphics[width=0.95\linewidth]{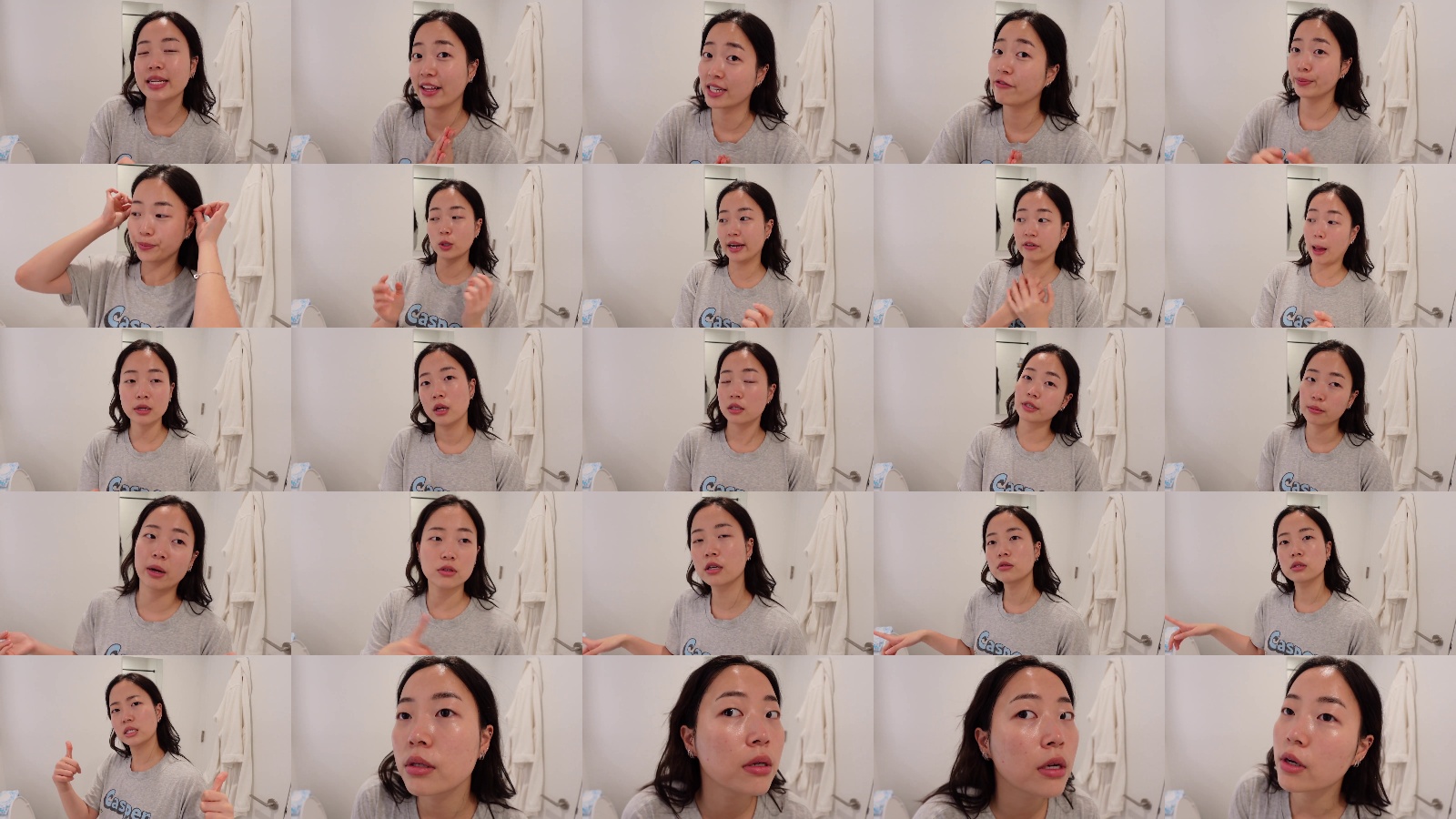}\\
\midrule
Good & \includegraphics[width=0.95\linewidth]{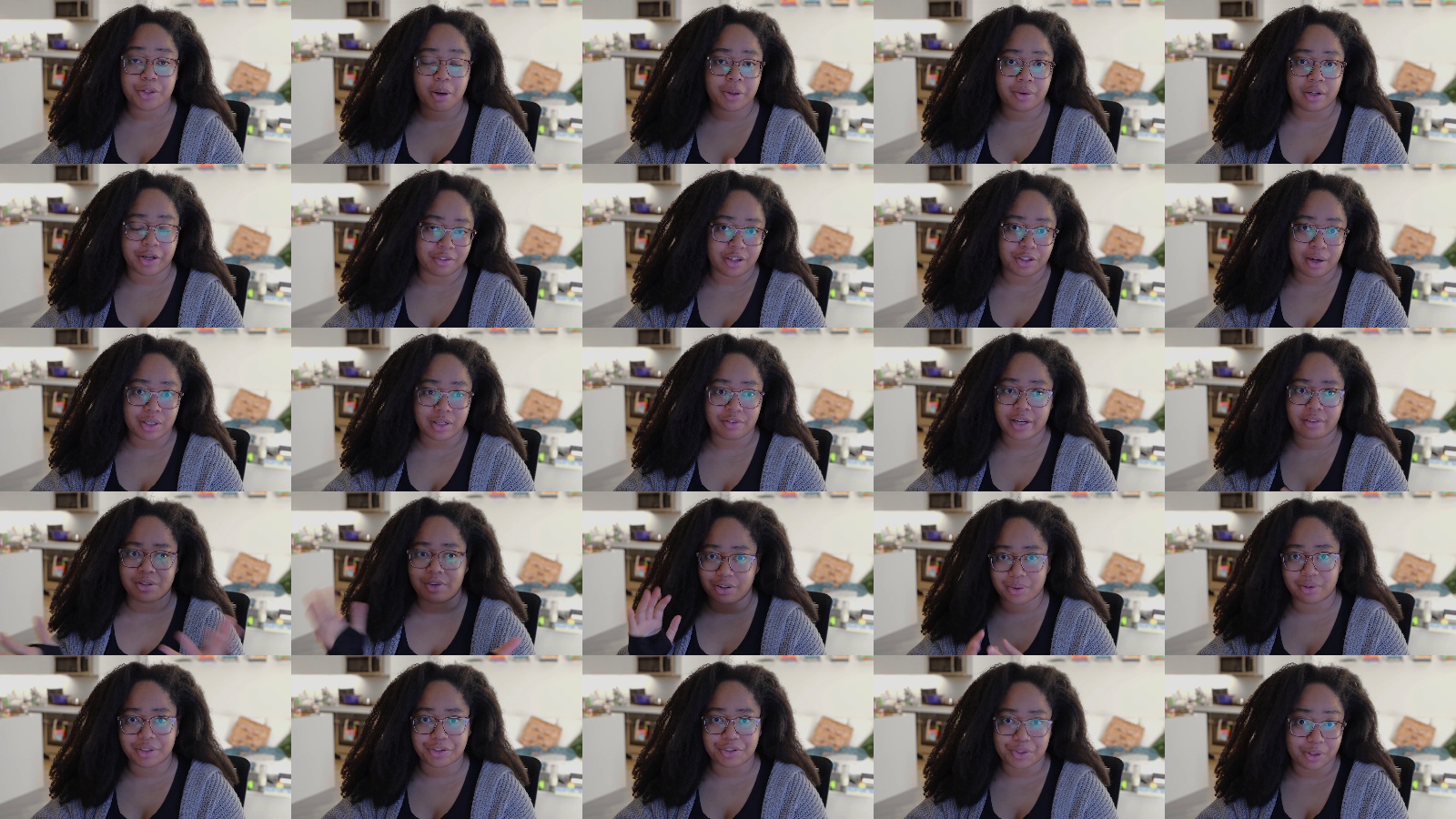}\\
\bottomrule
\end{tabular}
\caption{Examples of Sample Quality Based on Filter Cotracker}
\label{case_study_cotracker}
\end{table*}
\FloatBarrier

\begin{table*}[t]
\centering
\setlength{\tabcolsep}{6pt}      
\small
\begin{tabular}{@{}M{1.5cm}M{11.5cm}@{}}
\toprule
\multicolumn{2}{@{}>{\centering\arraybackslash}m{13cm}@{}}
{\textbf{Examples of Sample Quality Based on Filter Dover}}\\
\midrule
\textbf{Quality Level} & \textbf{Sample Example}\\
\midrule
Poor & \includegraphics[width=0.95\linewidth]{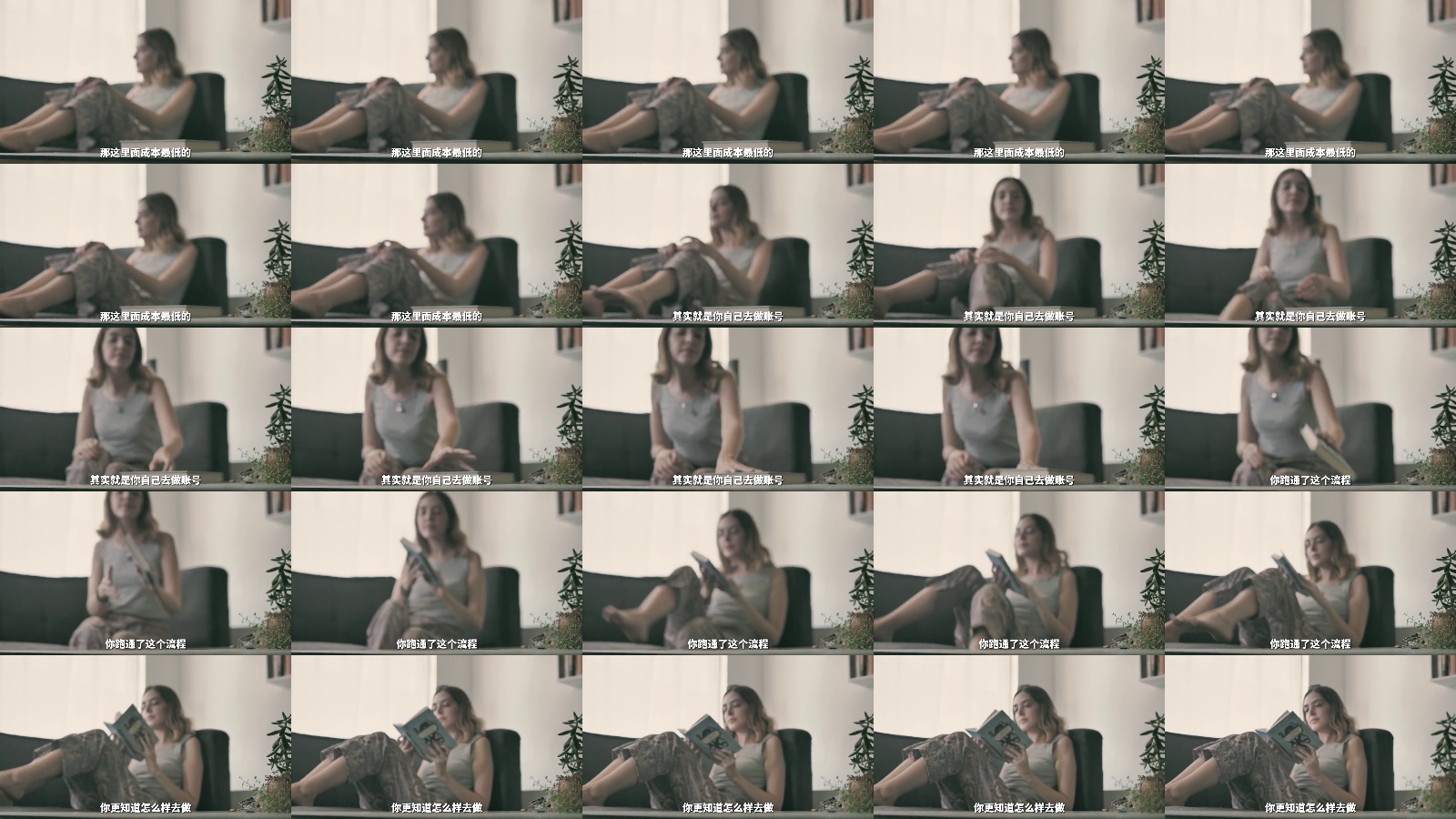}\\
\midrule
Fair & \includegraphics[width=0.95\linewidth]{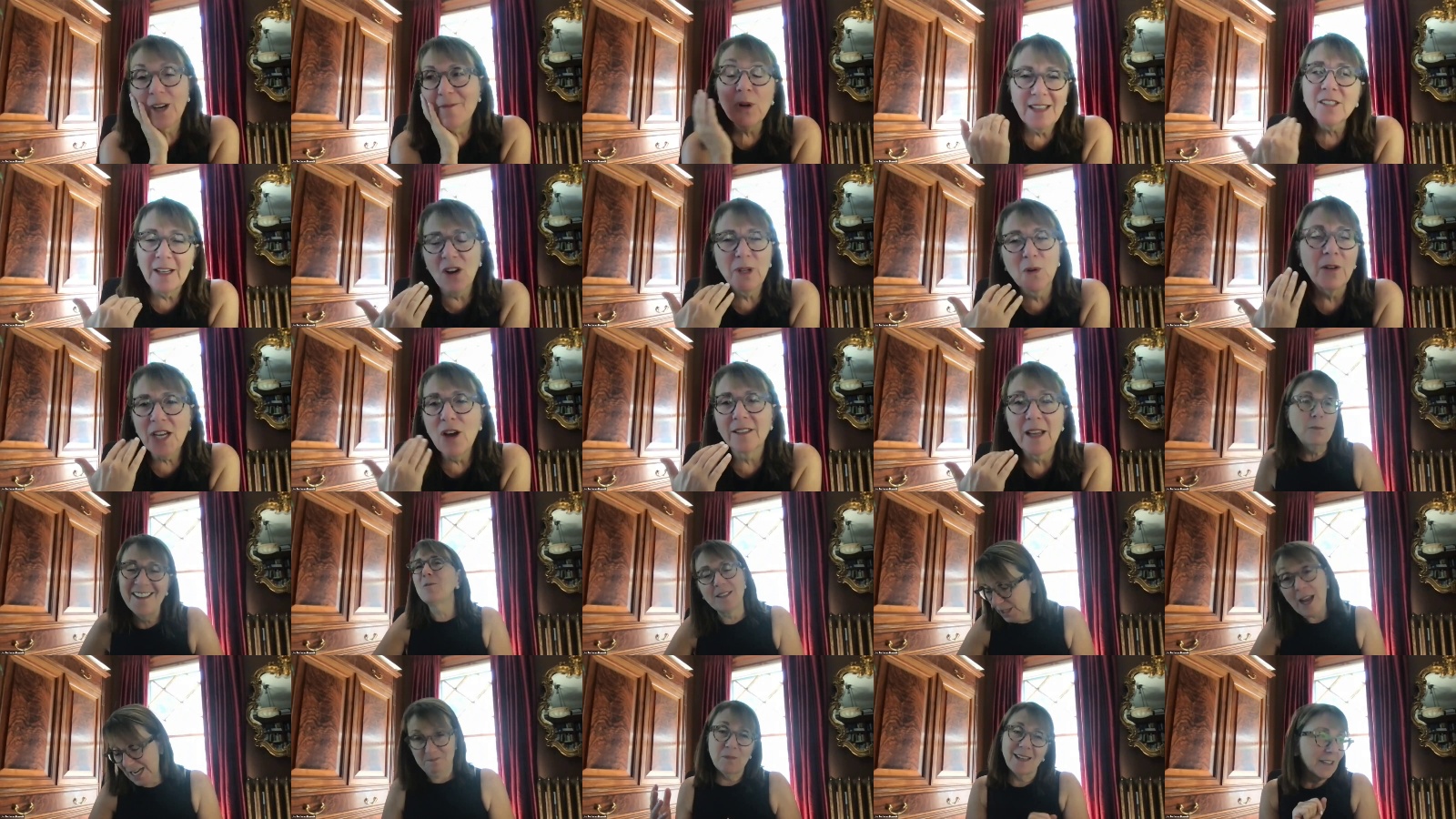}\\
\midrule
Good & \includegraphics[width=0.95\linewidth]{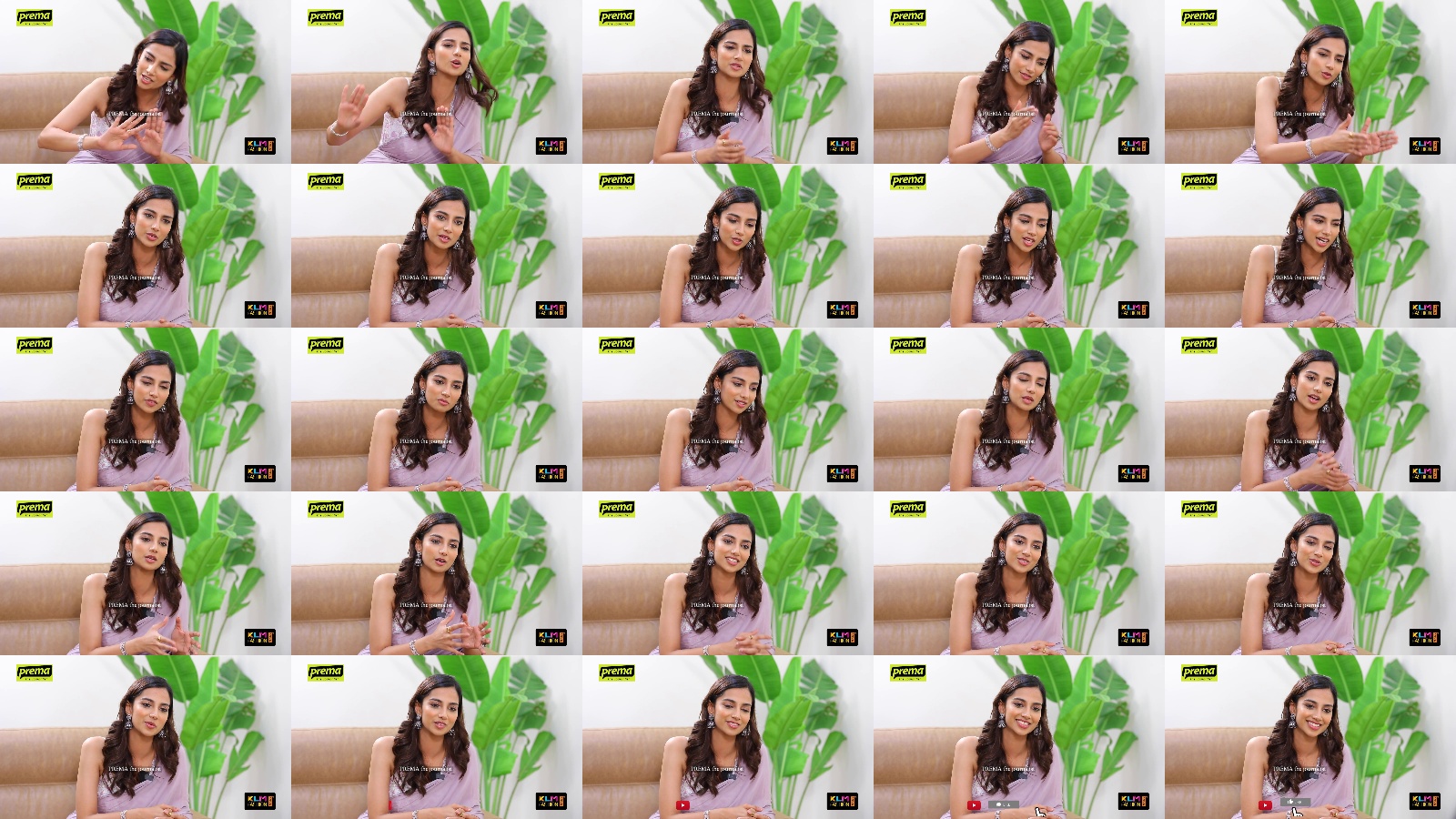}\\
\bottomrule
\end{tabular}%
\caption{Examples of Sample Quality Based on Filter Dover}
\label{case_study_dover}
\end{table*}
\FloatBarrier

\begin{table*}[t]
\centering
\setlength{\tabcolsep}{6pt}      
\small
\begin{tabular}{@{}M{1.5cm}M{11.5cm}@{}}
\toprule
\multicolumn{2}{@{}>{\centering\arraybackslash}m{13cm}@{}}
{\textbf{Examples of Sample Quality Based on Head Movement}}\\
\midrule
\textbf{Quality Level} & \textbf{Sample Example}\\
\midrule
Poor & \includegraphics[width=0.95\linewidth]{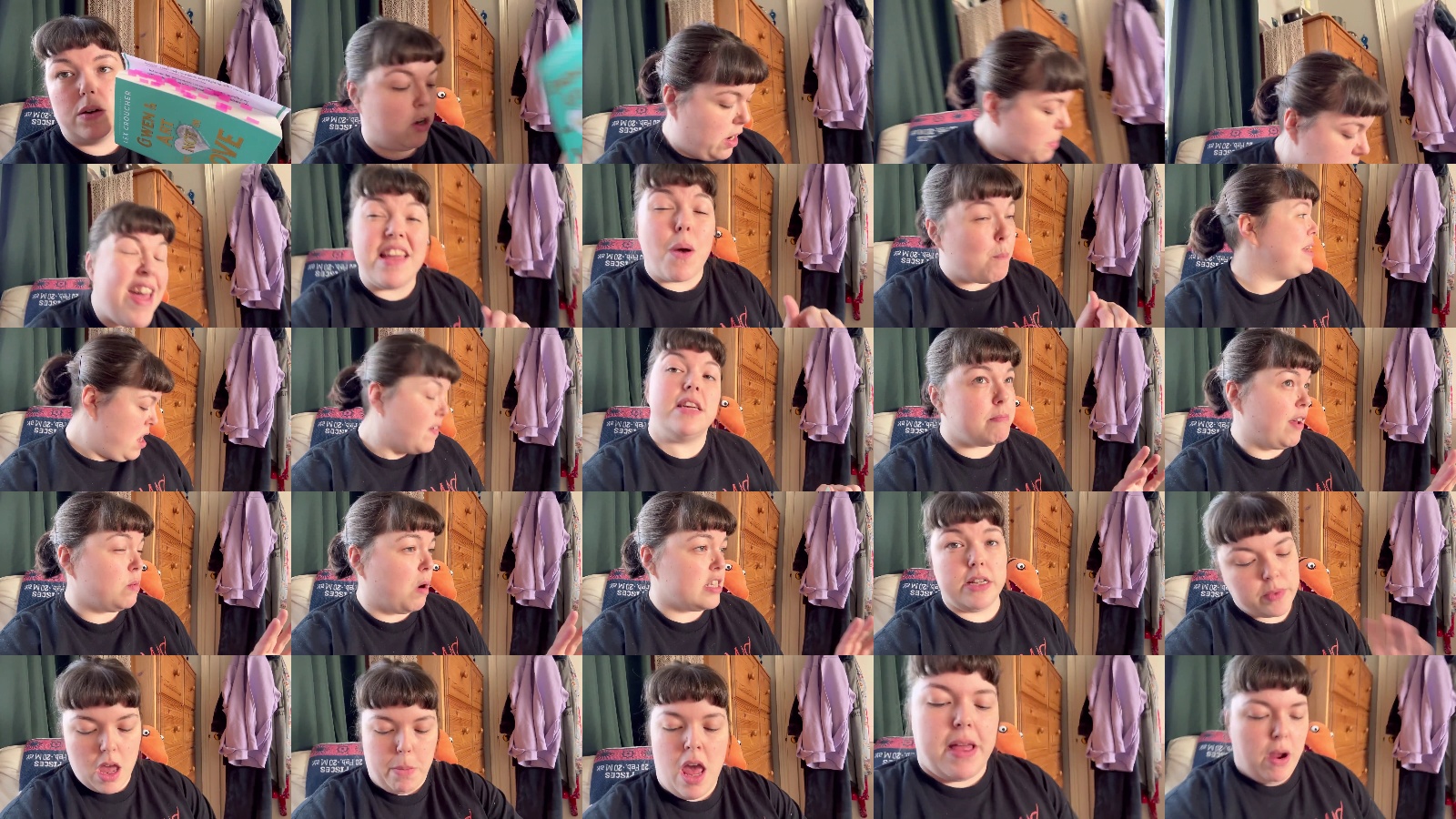}\\
\midrule
Fair & \includegraphics[width=0.95\linewidth]{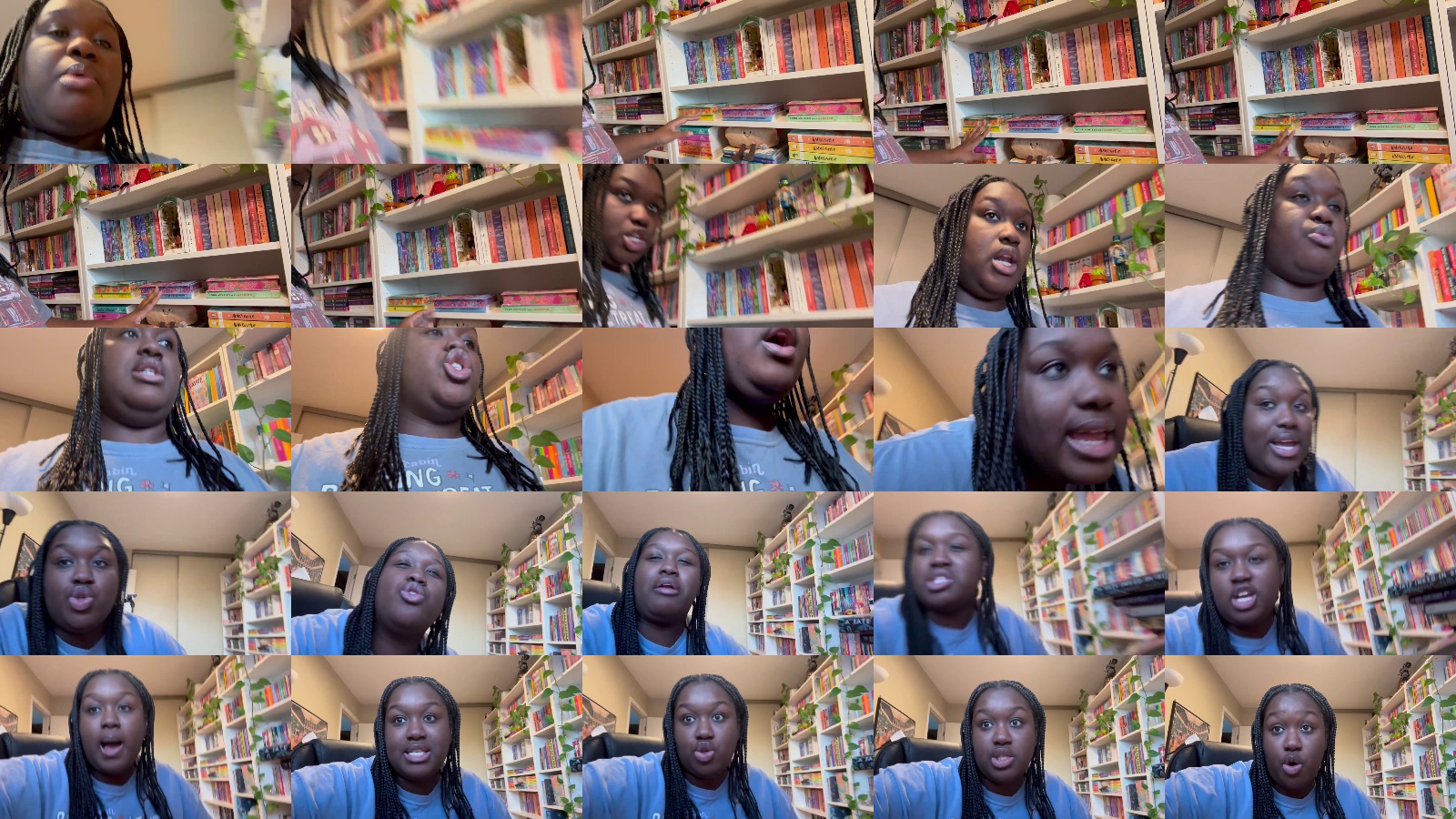}\\
\midrule
Good & \includegraphics[width=0.95\linewidth]{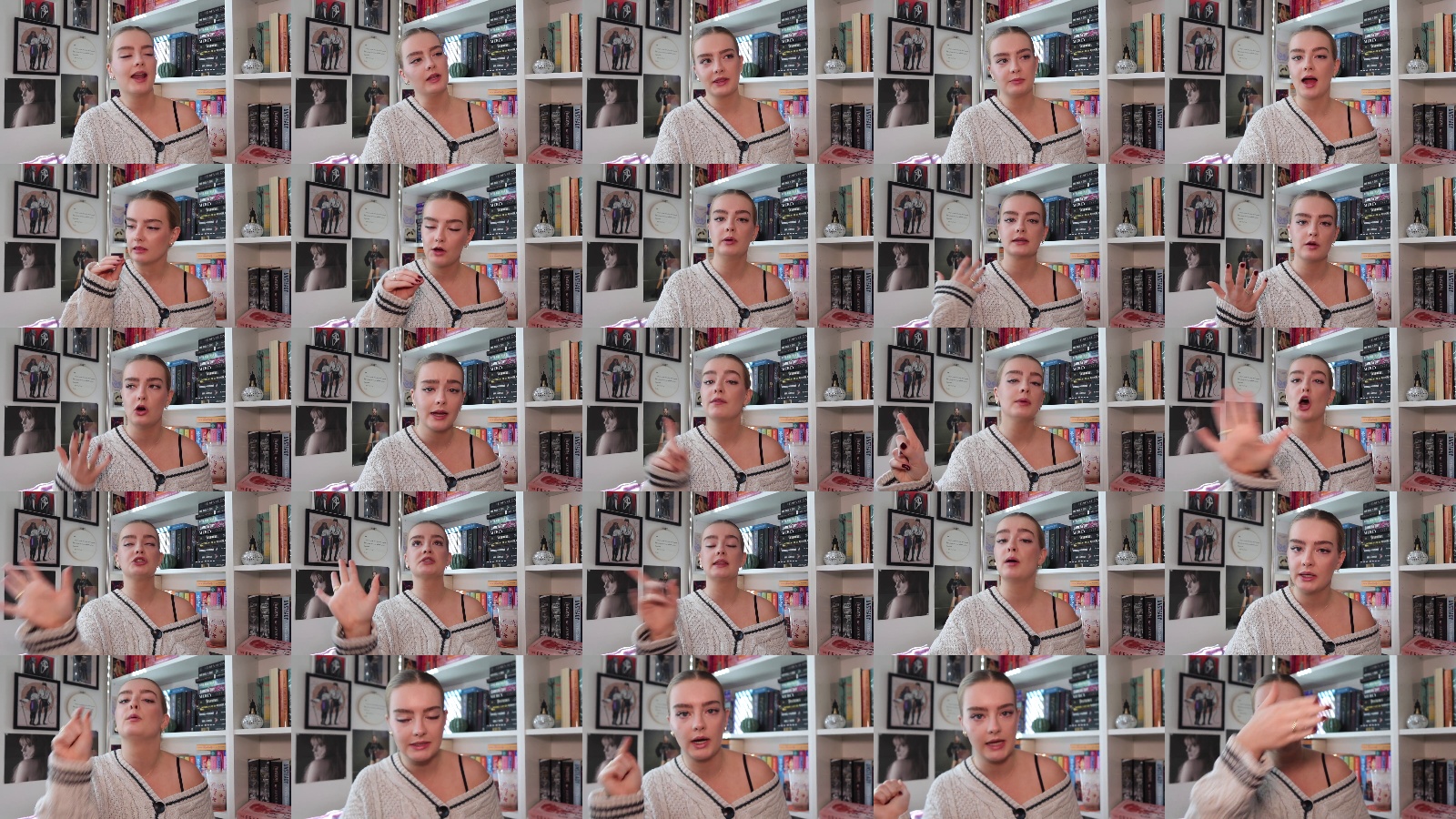}\\
\bottomrule
\end{tabular}%
\caption{Examples of Sample Quality Based on Filter Head Movement}
\label{case_study_head_movement}
\end{table*}
\FloatBarrier

\begin{table*}[t]
\centering
\setlength{\tabcolsep}{6pt}      
\small
\begin{tabular}{@{}M{1.5cm}M{11.5cm}@{}}
\toprule
\multicolumn{2}{@{}>{\centering\arraybackslash}m{13cm}@{}}
{\textbf{Examples of Sample Quality Based on Filter Head Orientation}}\\
\midrule
\textbf{Quality Level} & \textbf{Sample Example}\\
\midrule
Poor & \includegraphics[width=0.95\linewidth]{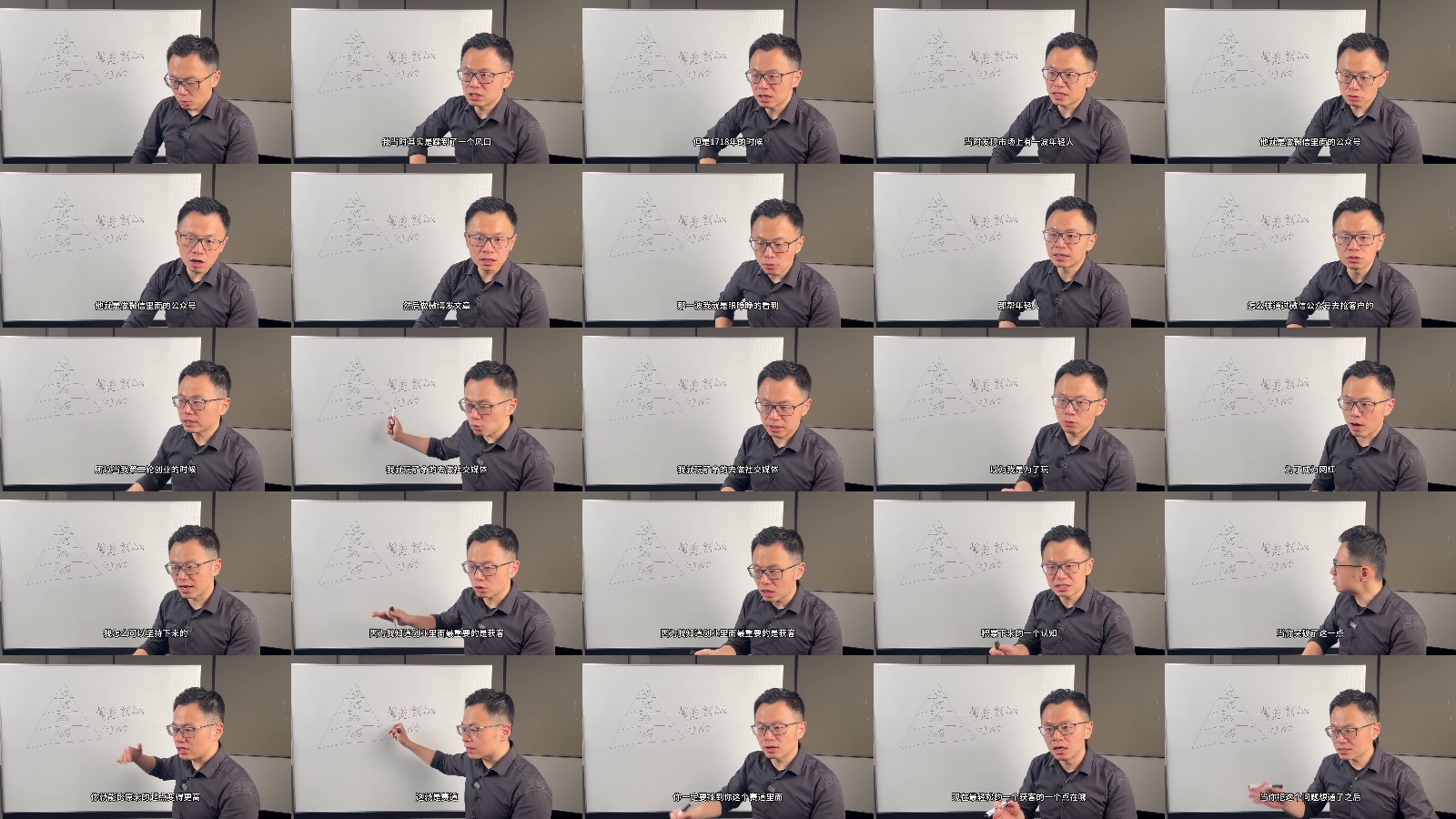}\\
\midrule
Fair & \includegraphics[width=0.95\linewidth]{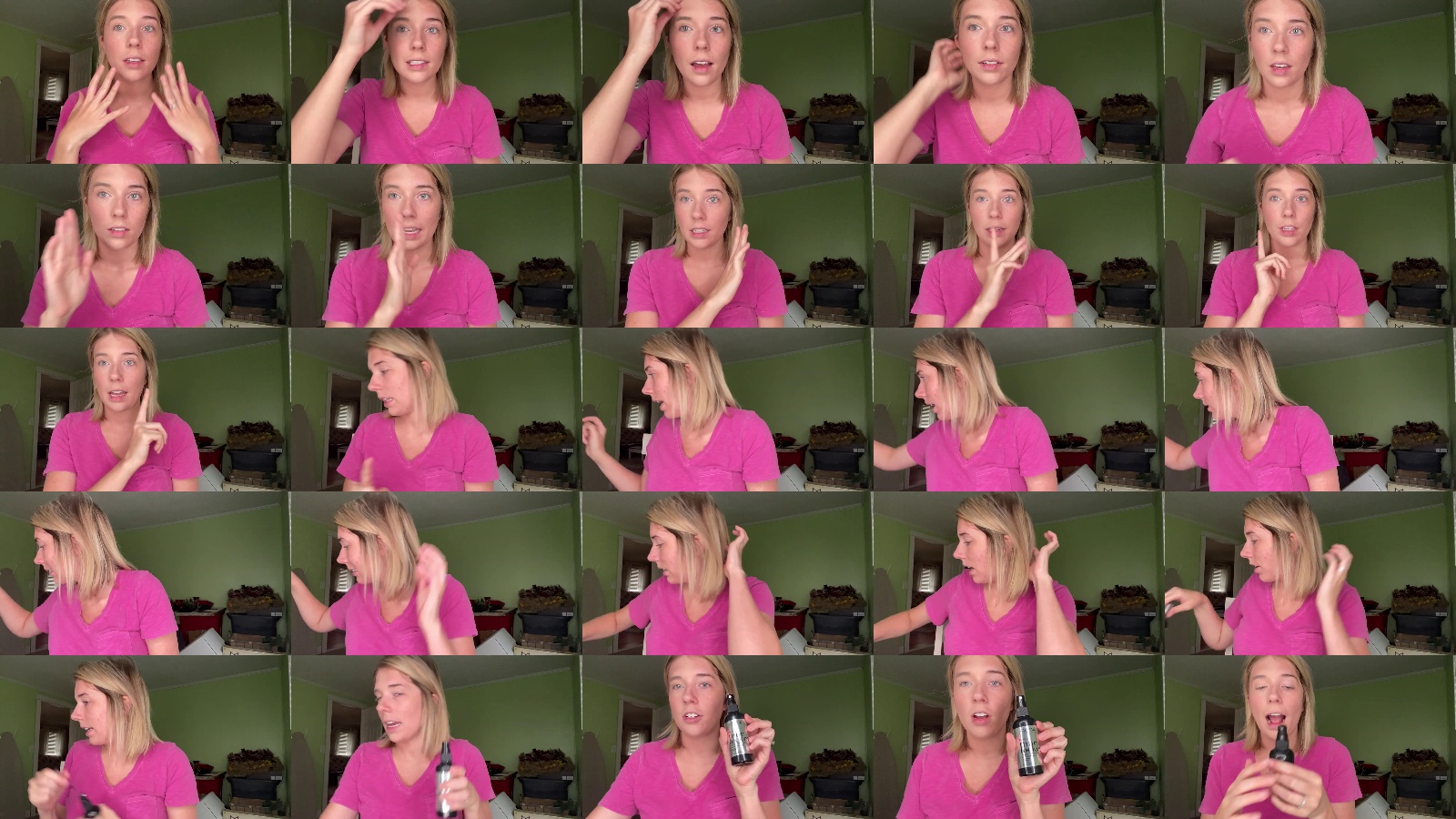}\\
\midrule
Good & \includegraphics[width=0.95\linewidth]{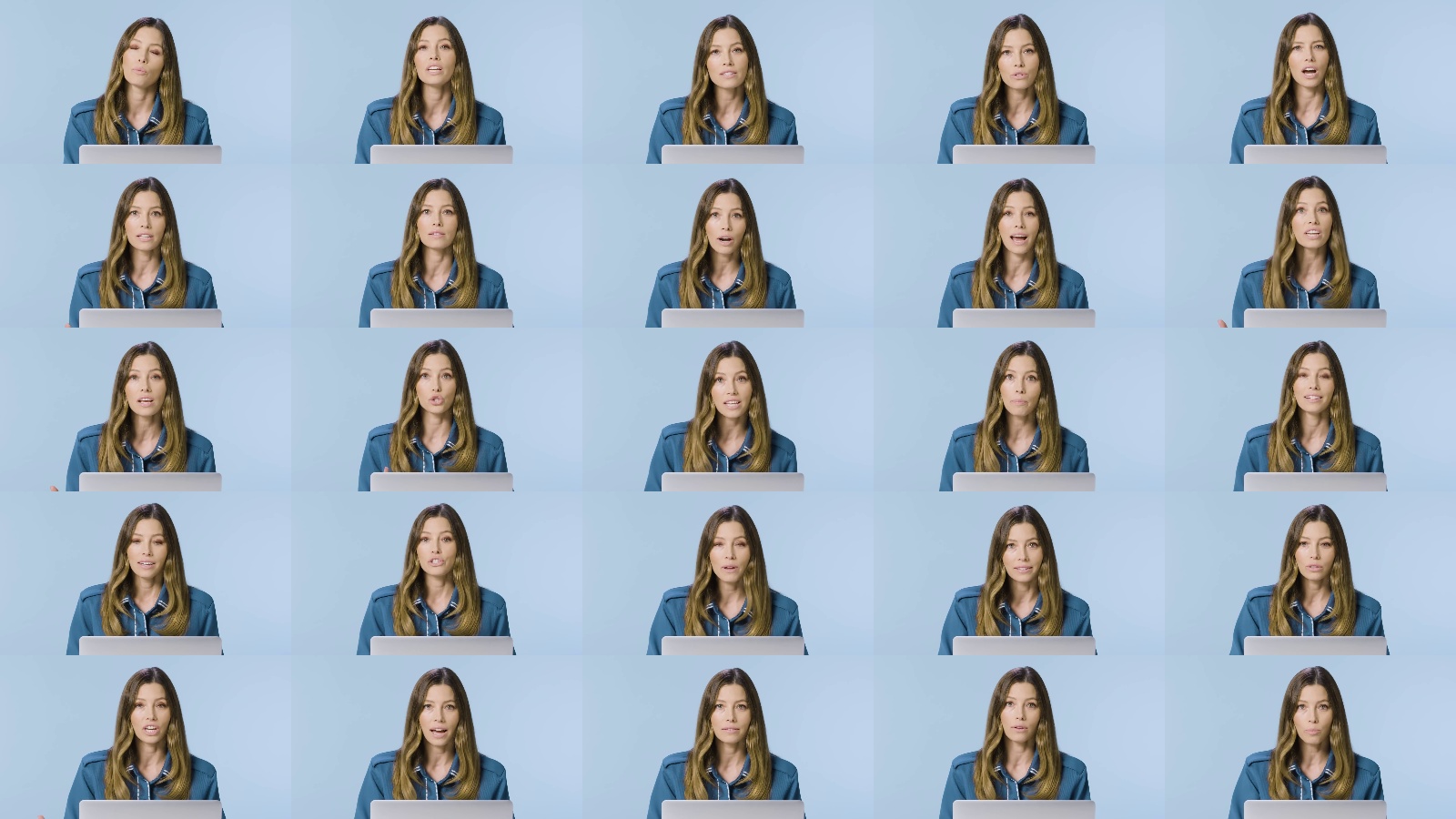}\\
\bottomrule
\end{tabular}
\caption{Examples of Sample Quality Based on Filter Head Orientation}
\label{case_study_head_orientation}
\end{table*}
\FloatBarrier

\begin{table*}[t]
\centering
\setlength{\tabcolsep}{6pt}      
\small
\begin{tabular}{@{}M{1.5cm}M{11.5cm}@{}}
\toprule
\multicolumn{2}{@{}>{\centering\arraybackslash}m{13cm}@{}}
{\textbf{Examples of Sample Quality Based on Filter Head Completeness}}\\
\midrule
\textbf{Quality Level} & \textbf{Sample Example}\\
\midrule
Poor & \includegraphics[width=0.95\linewidth]{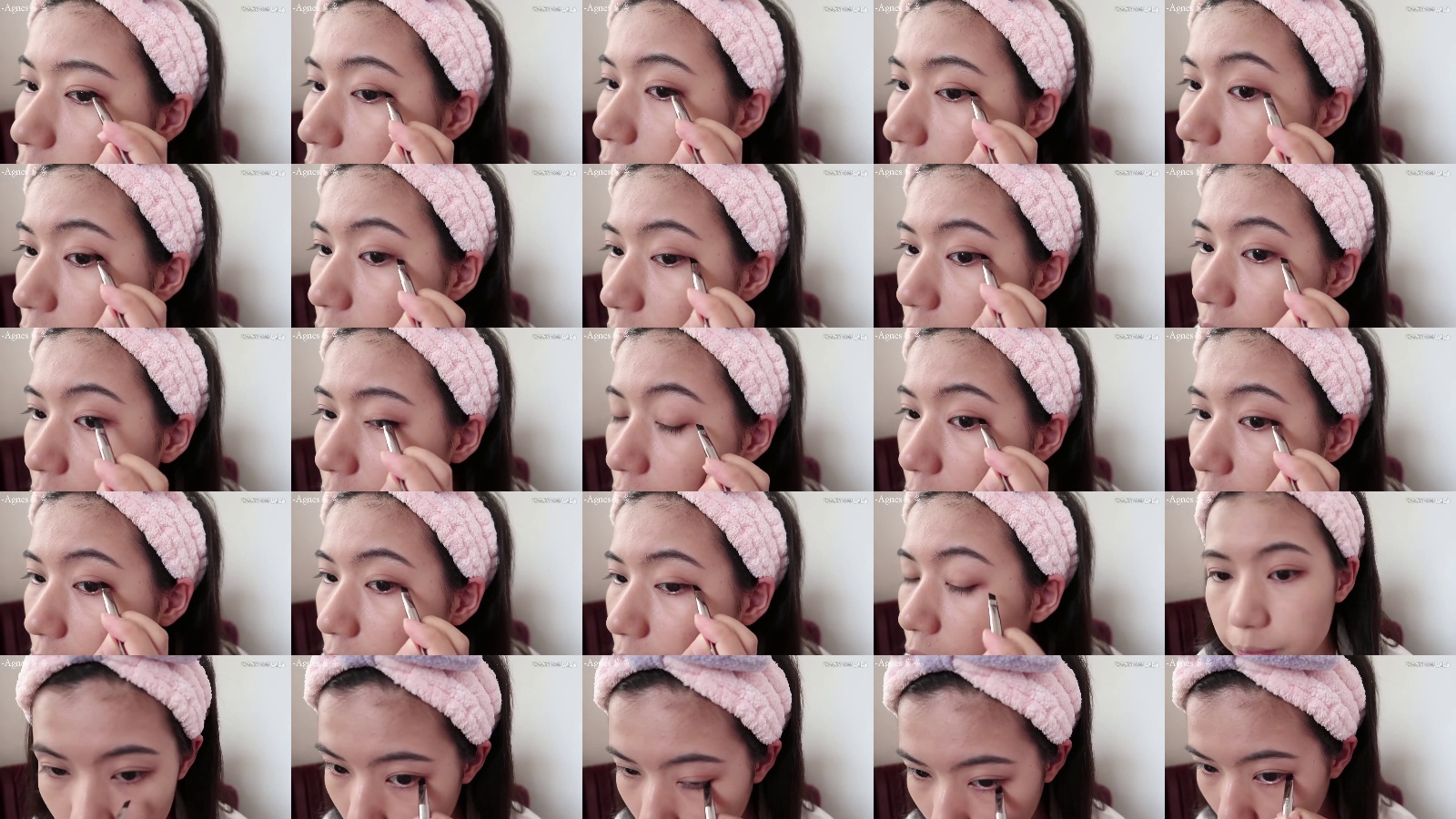}\\
\midrule
Fair & \includegraphics[width=0.95\linewidth]{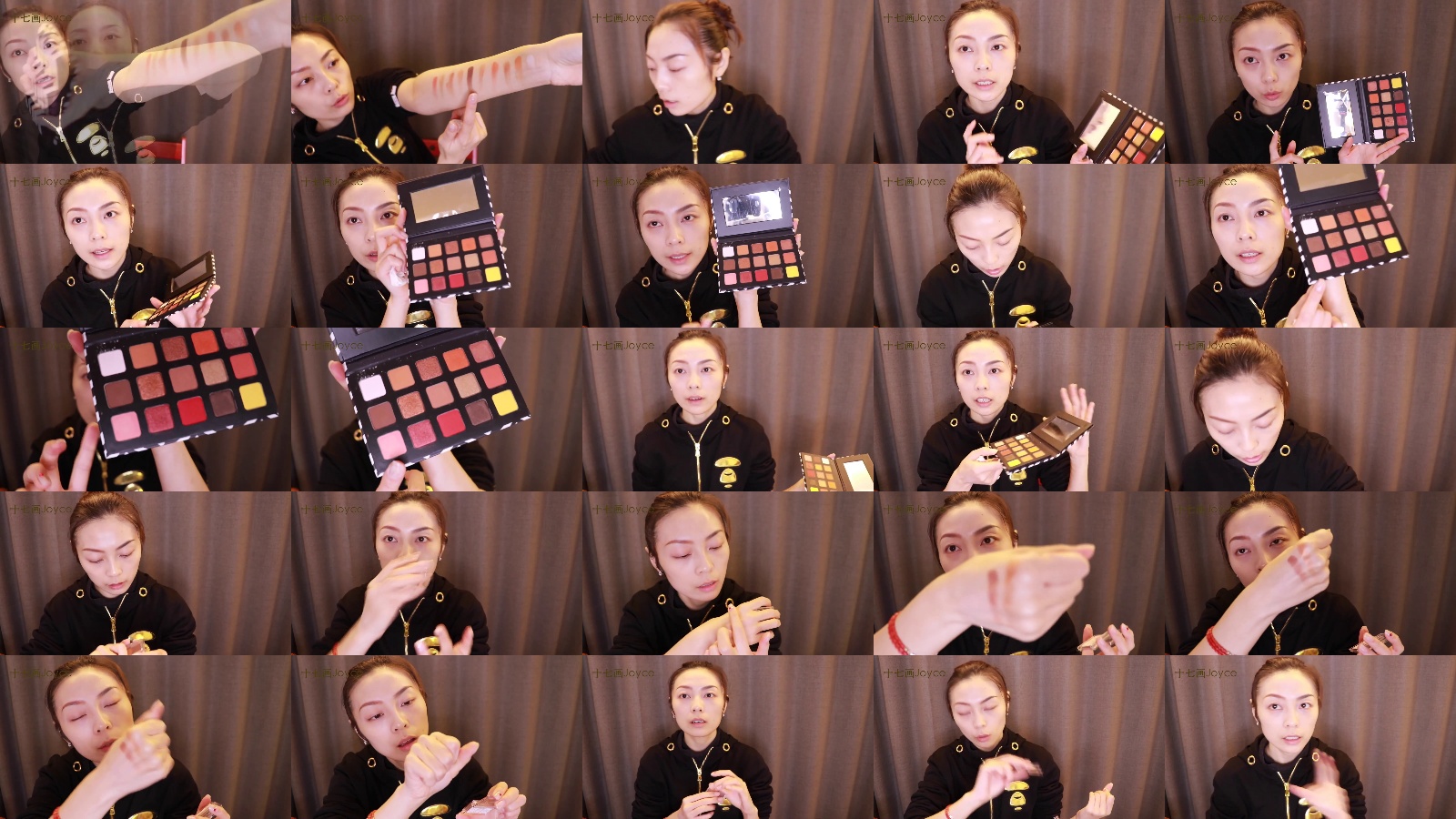}\\
\midrule
Good & \includegraphics[width=0.95\linewidth]{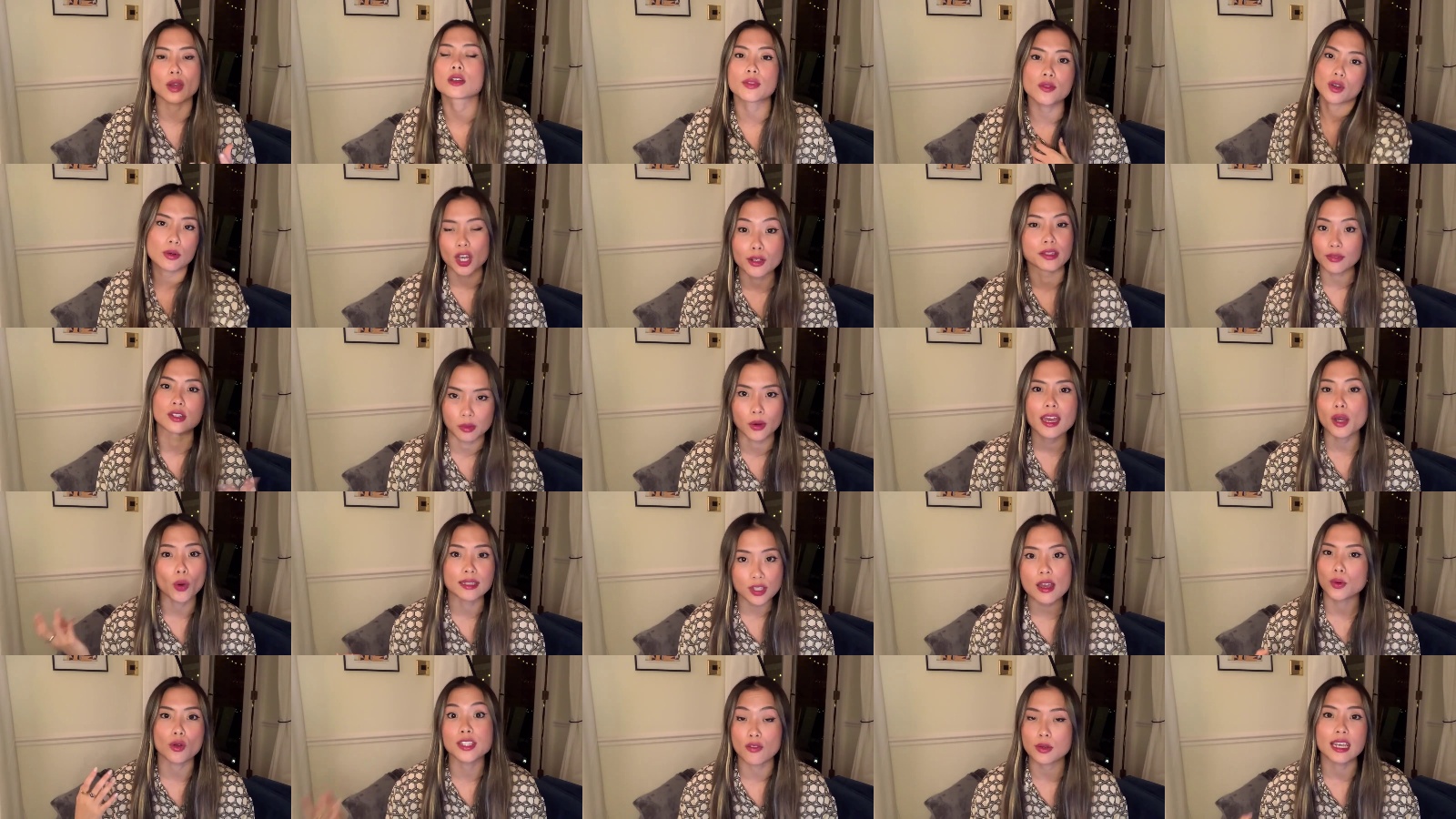}\\
\bottomrule
\end{tabular}
\caption{Examples of Sample Quality Based on Filter Head Completeness}
\label{case_study_head_completeness}
\end{table*}
\FloatBarrier

\begin{table*}[t]
\centering
\setlength{\tabcolsep}{6pt}      
\small
\begin{tabular}{@{}M{1.5cm}M{11.5cm}@{}}
\toprule
\multicolumn{2}{@{}>{\centering\arraybackslash}m{13cm}@{}}
{\textbf{Examples of Sample Quality Based on Filter Head Resolution}}\\
\midrule
\textbf{Quality Level} & \textbf{Sample Example}\\
\midrule
Poor & \includegraphics[width=0.95\linewidth]{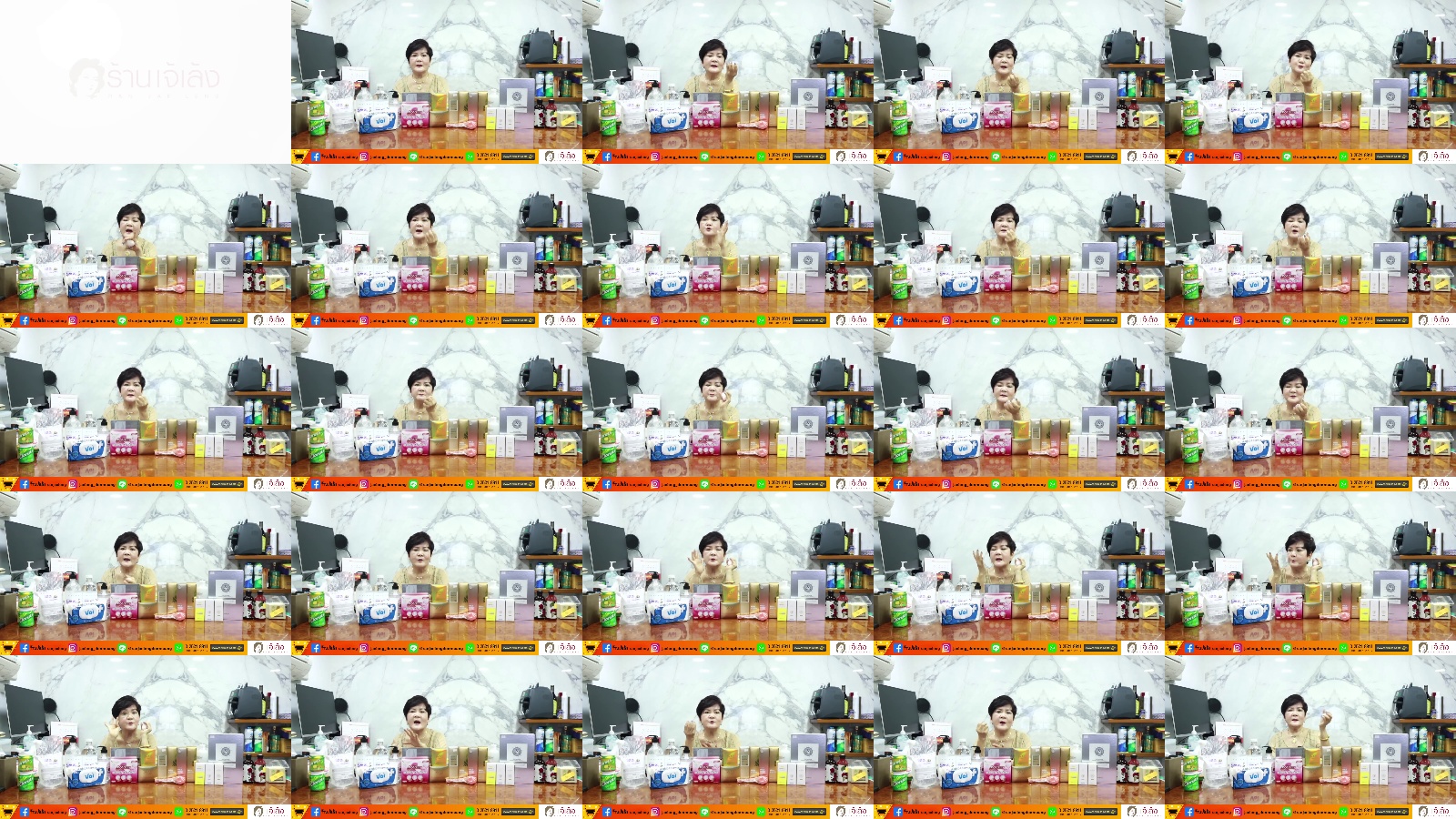}\\
\midrule
Fair & \includegraphics[width=0.95\linewidth]{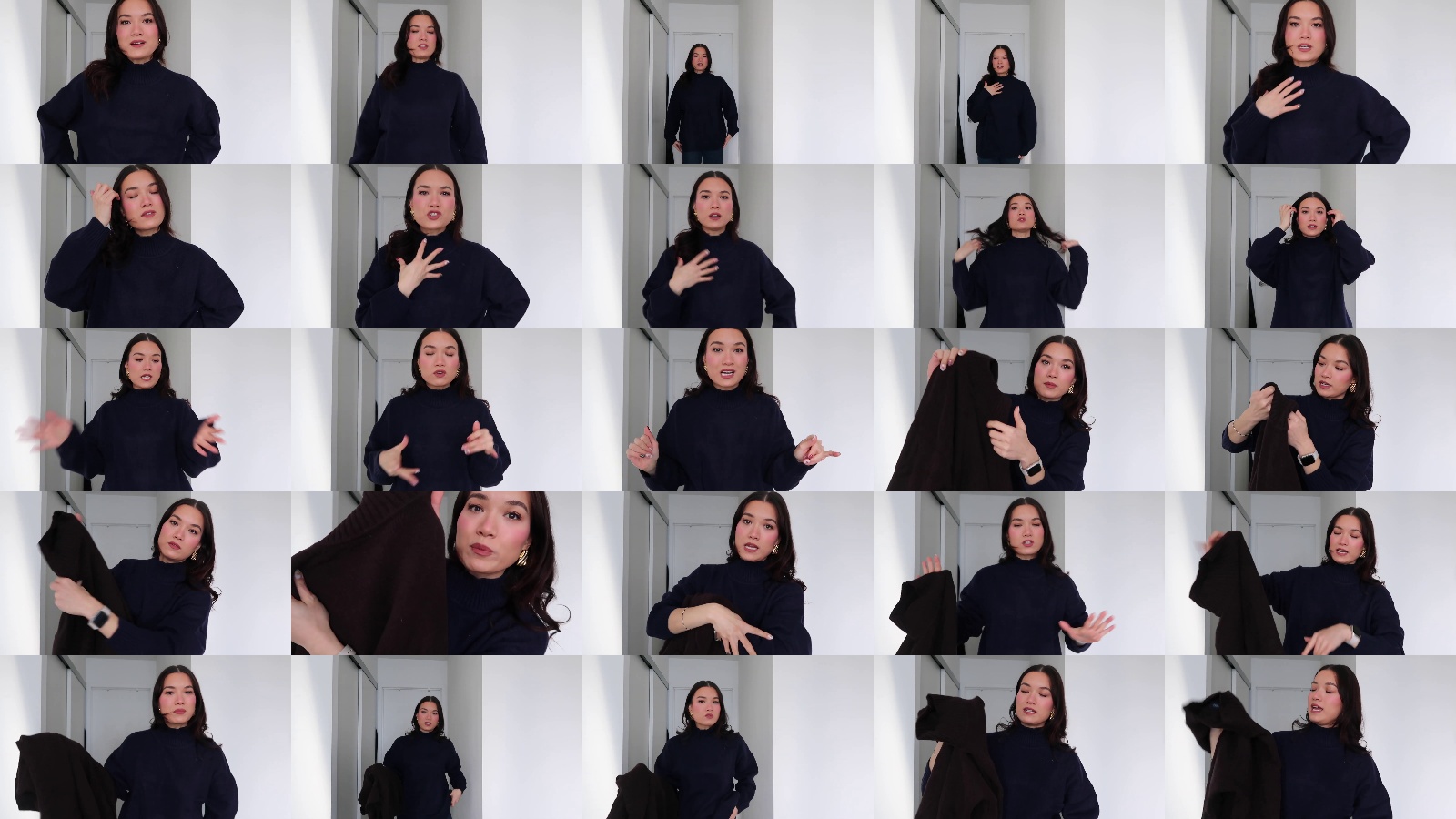}\\
\midrule
Good & \includegraphics[width=0.95\linewidth]{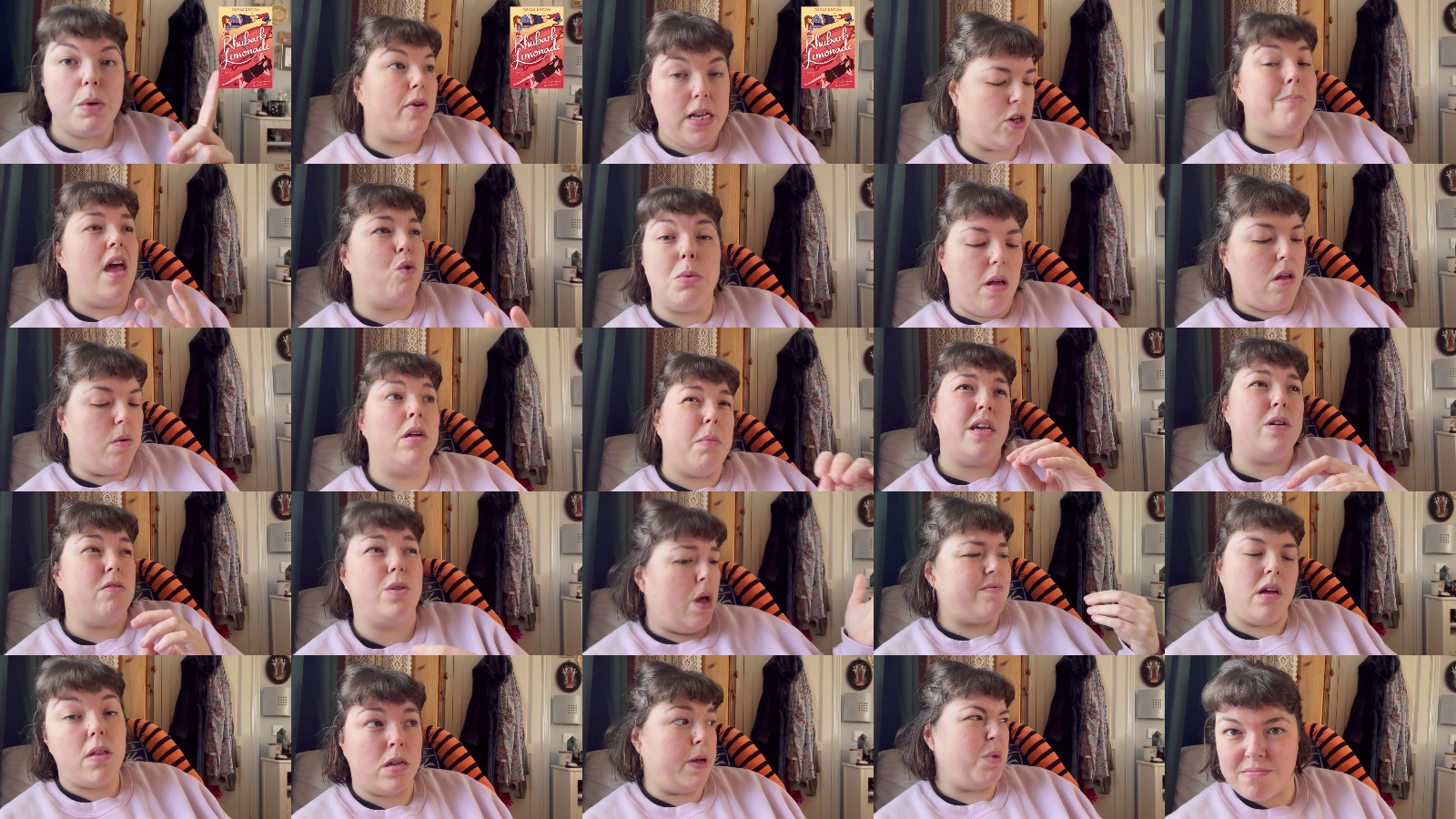}\\
\bottomrule
\end{tabular}
\caption{Examples of Sample Quality Based on Filter Head Resolution}
\label{case_study_head_resolution}
\end{table*}
\FloatBarrier

\begin{table*}[t]
\centering
\setlength{\tabcolsep}{6pt}      
\small
\begin{tabular}{@{}M{1.5cm}M{11.5cm}@{}}
\toprule
\multicolumn{2}{@{}>{\centering\arraybackslash}m{13cm}@{}}
{\textbf{Examples of Sample Quality Based on Filter Head Rotation}}\\
\midrule
\textbf{Quality Level} & \textbf{Sample Example}\\
\midrule
Poor & \includegraphics[width=0.95\linewidth]{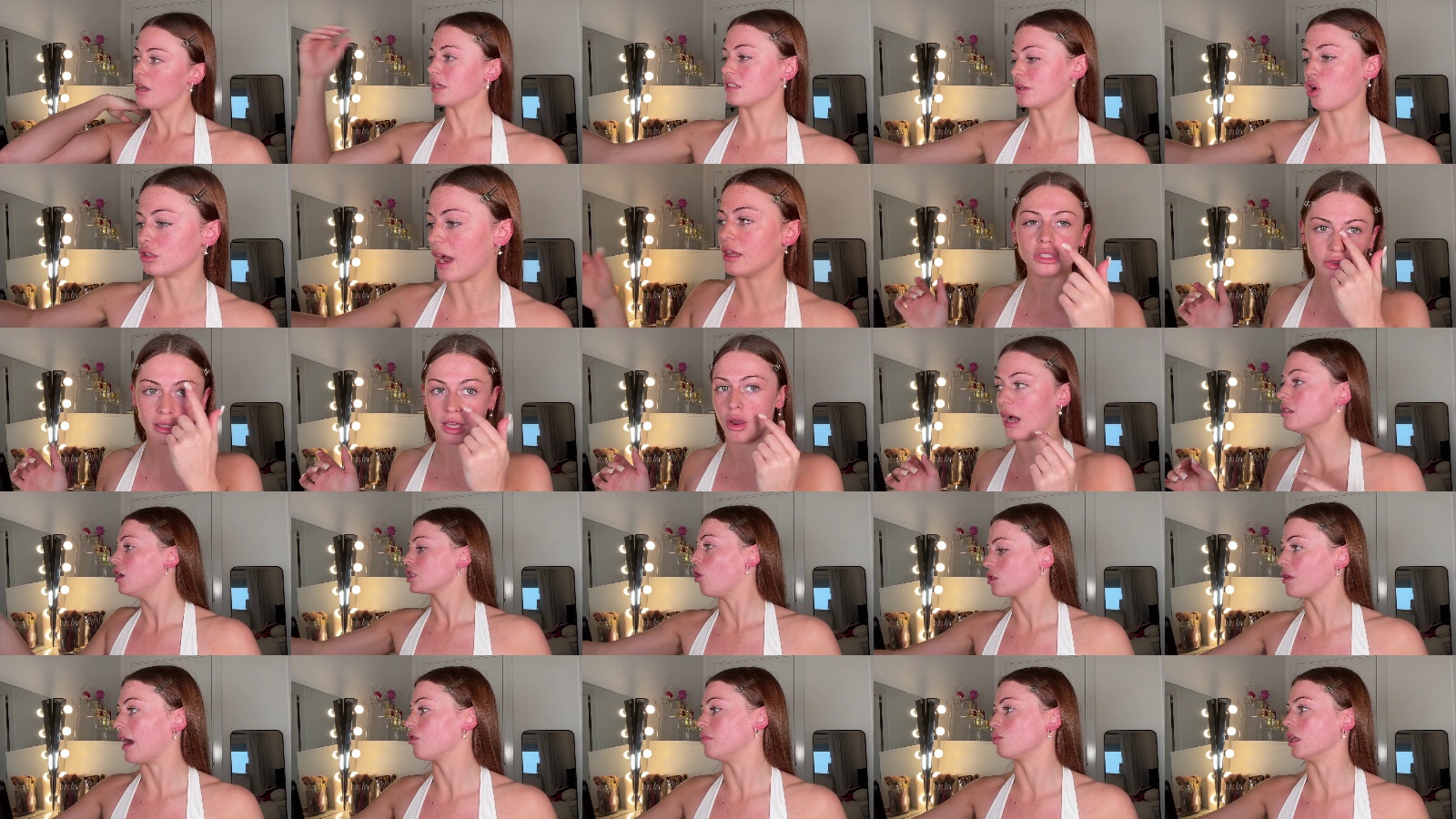}\\
\midrule
Fair & \includegraphics[width=0.95\linewidth]{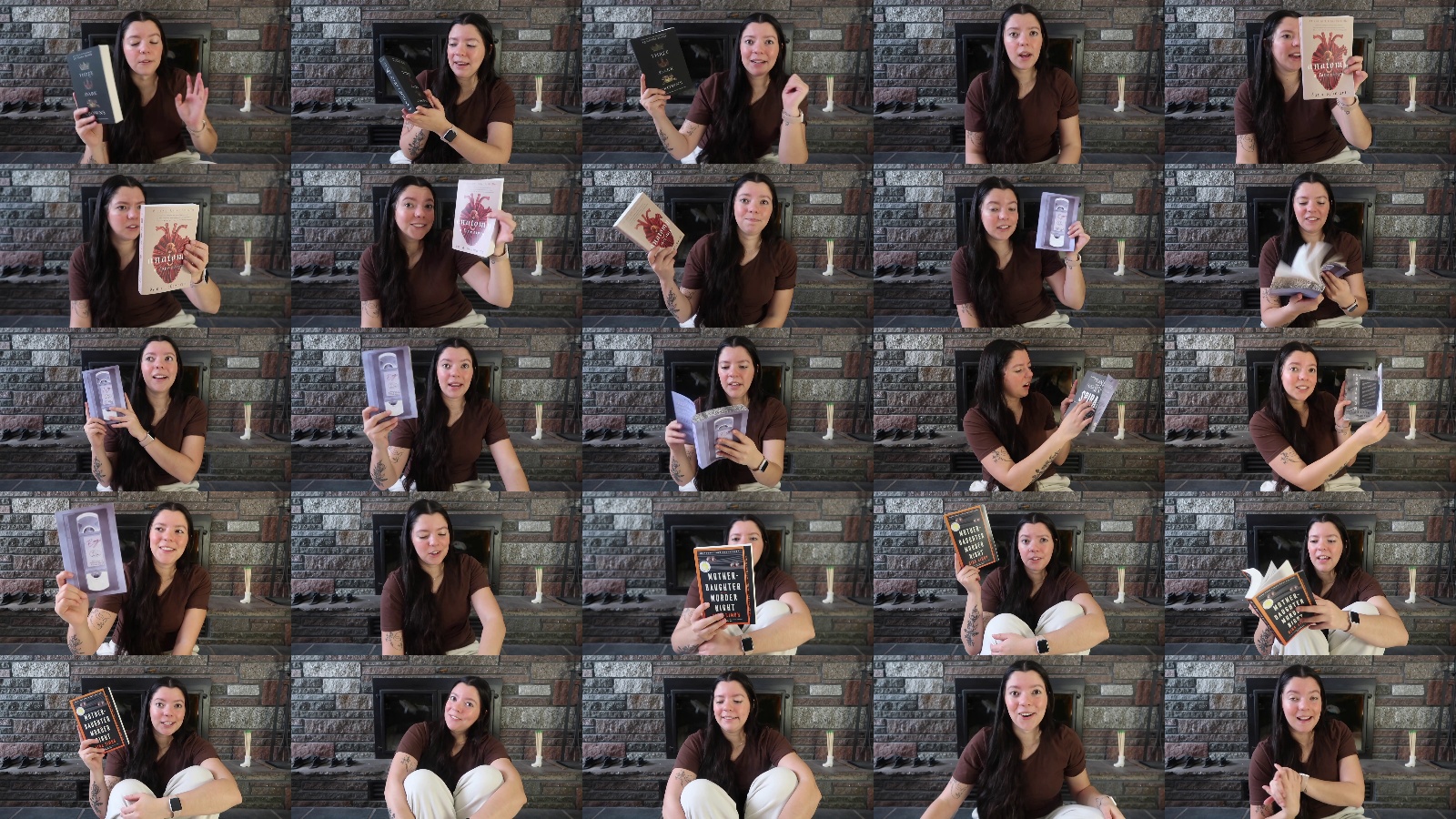}\\
\midrule
Good & \includegraphics[width=0.95\linewidth]{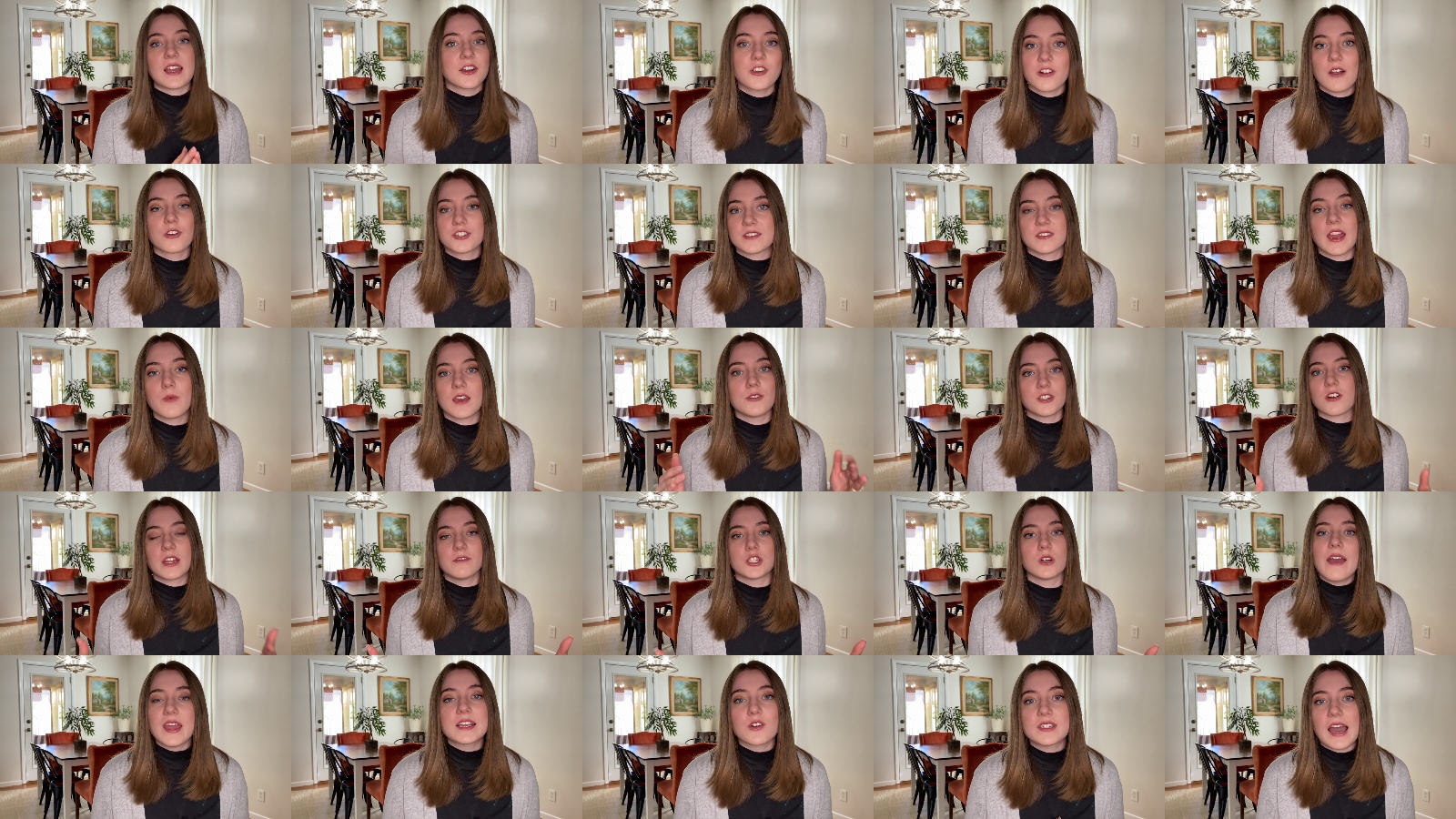}\\
\bottomrule
\end{tabular}
\caption{Examples of Sample Quality Based on Filter Head Rotation}
\label{case_study_head_rotation}
\end{table*}
\FloatBarrier

\section{Annotation Visualization}
\label{app:annotation_visualization}
This section provides a visual overview of the generated annotations for the TalkVid-Core dataset.

\subsection{Annotation Details}
We prompt the model to perform a detailed analysis focusing on anatomical movement patterns and behavioral dynamics, returning its findings as a structured JSON object. This process yields a rich set of 180,860 structured annotations. To quantify their richness, the descriptions have an average length of \textbf{84.6 tokens} ($\sigma=28.4$) and draw from a diverse vocabulary of over \textbf{18,000 unique tokens}, underscoring the detail and variety of the generated text. 

\subsection{Annotation Length Distribution}

\begin{figure}[ht!]
    \centering
    \includegraphics[width=0.8\linewidth]{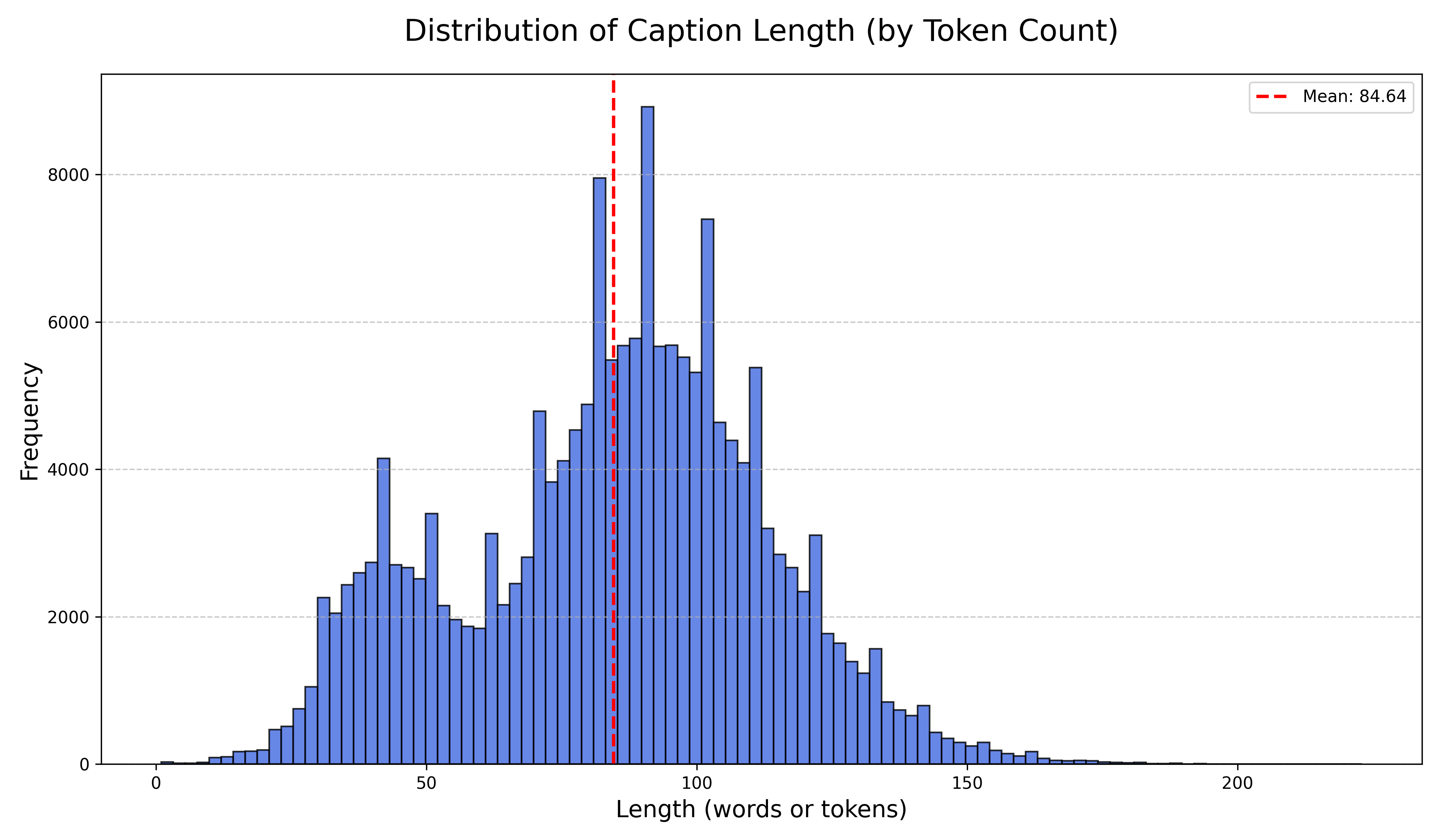}
    \caption{Distribution of caption lengths by token count across all 180k annotations in TalkVid-Core. The mean length is 84.6 tokens, with a standard deviation of 28.4, indicating a consistent descriptive depth.}
\end{figure}

\subsection{Qualitative Examples} 

\lstdefinestyle{promptstyle}{
    backgroundcolor=\color{black!5},   
    basicstyle=\ttfamily\footnotesize, 
    breaklines=true,                   
    frame=tb,                          
    captionpos=b,
    keepspaces=true,
    showstringspaces=false,
}

\lstdefinestyle{jsonstyle}{
    backgroundcolor=\color{black!5},
    basicstyle=\ttfamily\footnotesize,
    breaklines=true,
    frame=single,                      
    captionpos=b,
    keepspaces=true,
    stringstyle=\color{purple!80!black},
    keywordstyle=\color{blue},           
    morekeywords={true,false,null},
    showstringspaces=false,
}

\section{Details of Annotation Generation Process}
\label{app:generation_details} 

This section provides the full prompt used for annotation generation and a representative example of the structured JSON output.

\subsection{Full Generation Prompt}
\label{app:full_prompt}

The following prompt was provided to a large multimodal model to instruct it on the analysis task and the required output format.

\begin{lstlisting}[style=promptstyle]
Analyze the video with precise focus on anatomical movement patterns and behavioral dynamics. Prioritize detailed descriptions of body part trajectories and their temporal relationships.

IMPORTANT OUTPUT REQUIREMENTS:
1. Provide your analysis in a clean, minified JSON format without any line breaks or escape characters
2. Do not include any explanatory text before or after the JSON
3. Ensure the JSON is valid and properly formatted
4. Use single-line format for the entire output
5. Do not include any comments or additional formatting

Expected JSON structure:
{
  "scene_context": {
    "background": ["Setting description", "Environmental elements", "Lighting details", "Camera angle and position"],
    "subject": ["Physical appearance", "Clothing description", "Notable features", "Demographic attributes"]
  },
  "movement_analysis": {
    "head_movements": ["Sequential head rotations with angle measurements (in degrees)", "Tilt progression with directional markers", "Orientation changes relative to initial position", "Facial expressions and changes"],
    "hand_actions": {
      "left": ["Trajectory patterns with spatial coordinates", "Gesture type classification", "Interaction duration and objects", "Force/intensity of movements"],
      "right": ["Trajectory patterns with spatial coordinates", "Gesture type classification", "Interaction duration and objects", "Force/intensity of movements"]
    },
    "torso_movements": ["Rotation angles and direction", "Flexion/extension patterns", "Lateral movements", "Weight distribution shifts", "Postural stability assessment"],
    "body_posture": ["Postural transition timeline", "Spinal alignment changes", "Weight distribution shifts", "Balance and stability patterns"],
    "confidence_scores": {
      "head_tracking": 0.0,
      "hand_tracking": 0.0,
      "torso_tracking": 0.0,
      "posture_analysis": 0.0,
      "overall_confidence": 0.0
    }
  },
  "interactions": {
    "environmental": ["Object interactions", "Space utilization", "Environmental adaptations"],
    "social": ["Interpersonal distances", "Social gestures", "Interactive behaviors"],
    "object_handling": ["Object types", "Manipulation patterns", "Duration of interactions"]
  },
  "audio_behavioral_analysis": {
    "speech": ["Voice characteristics", "Speech patterns", "Verbal expressions"],
    "non_verbal_sounds": ["Types of sounds", "Timing", "Context"]
  },
  "movement_metrics": {
    "speed": {"unit": "meters/second", "measurements": []},
    "acceleration": {"unit": "meters/second^2", "measurements": []},
    "angular_velocity": {"unit": "degrees/second", "measurements": []}
  },
  "anomaly_detection": {
    "unusual_movements": [],
    "irregular_patterns": [],
    "potential_concerns": []
  },
  "description_summary": "Comprehensive summary describing the complete action sequence"
}

RESPONSE FORMAT:
- Output must be a single JSON object without any additional text
- Do not include any markdown formatting
- Do not include any explanatory text
- Provide direct values without any placeholder text
- The entire response should be valid JSON that can be parsed directly
\end{lstlisting}

\subsection{Example of Generated JSON Output}
\label{app:json_example}

This is a representative example of the structured JSON data produced by the model for a single video clip.

\begin{lstlisting}[style=jsonstyle]
{
  "scene_context": {
    "background": [
      "Indoor setting",
      "Plain, light-colored wall",
      "Even, bright lighting"
    ],
    "subject": [
      "Middle-aged woman, light skin tone",
      "Dark pink/maroon long-sleeved top",
      "Shoulder-length blonde hair",
      "Presumably Caucasian"
    ]
  },
  "movement_analysis": {
    "head_movements": [
      "Slight head nodding throughout the video, particularly emphasizing certain words.  Vertical rotation within approximately 5 degrees.",
      "Minimal head tilt, maintaining a neutral position relative to the camera."
    ],
    "hand_actions": {
      "left": [
        "Initially rests at her side, out of frame.",
        "At approximately 0:06 seconds, her left hand rises into the frame and executes a pointing gesture, moving upward in the sagittal plane and extending forward in the frontal plane.",
        "The hand remains raised, with slight movements emphasizing speech."
      ],
      "right": [
        "Initially at her side and enters the frame slightly before the left hand, around 0:05 seconds.",
        "Performs a similar pointing gesture as the left hand, rising in the sagittal plane and extending forward in the frontal plane, but with more pronounced movement.",
        "The right hand is more active, continuing to gesture with varied movements in all three planes (sagittal, frontal, and transverse) throughout the video to emphasize points."
      ]
    },
    "body_posture": [
      "The subject maintains a stationary, standing posture throughout the video.",
      "Upright posture with minimal spinal curvature changes.",
      "Shoulders remain relaxed and relatively still.",
      "Weight distribution appears evenly balanced."
    ]
  },
  "description_summary": "The video shows a woman standing against a plain background, delivering a short explanation.  She wears a dark pink top. From a medium close-up shot framing her from the chest up, she speaks directly to the camera. Throughout the video, her head remains relatively still with subtle nodding for emphasis. Her hands, initially at her sides, rise into the frame to perform illustrative pointing gestures. The right hand exhibits more dynamic movement, emphasizing key points with a variety of gestures in multiple planes. Her overall body posture remains static, standing upright and facing forward, with weight evenly distributed. Her torso and shoulders show minimal movement."
}
\end{lstlisting}

\begin{figure*}[t!]
    \centering
    \begin{subfigure}[t]{0.32\linewidth}
        \includegraphics[width=\linewidth]{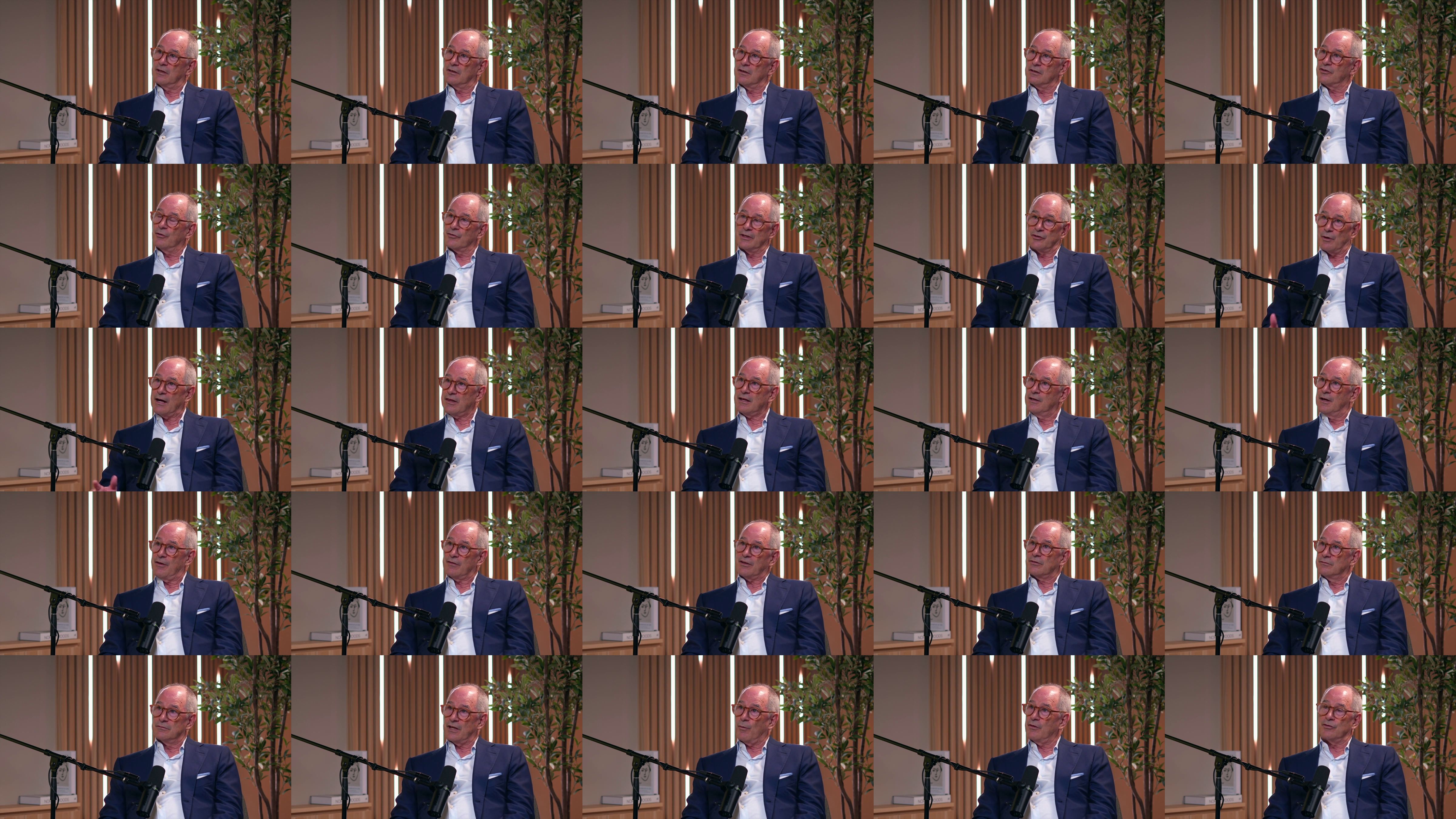}
        \caption*{\footnotesize The image collage shows a man sitting in a studio setting... He maintains a consistent seated posture... no significant body movement is observed.}
    \end{subfigure}
    \hfill
    \begin{subfigure}[t]{0.32\linewidth}
        \includegraphics[width=\linewidth]{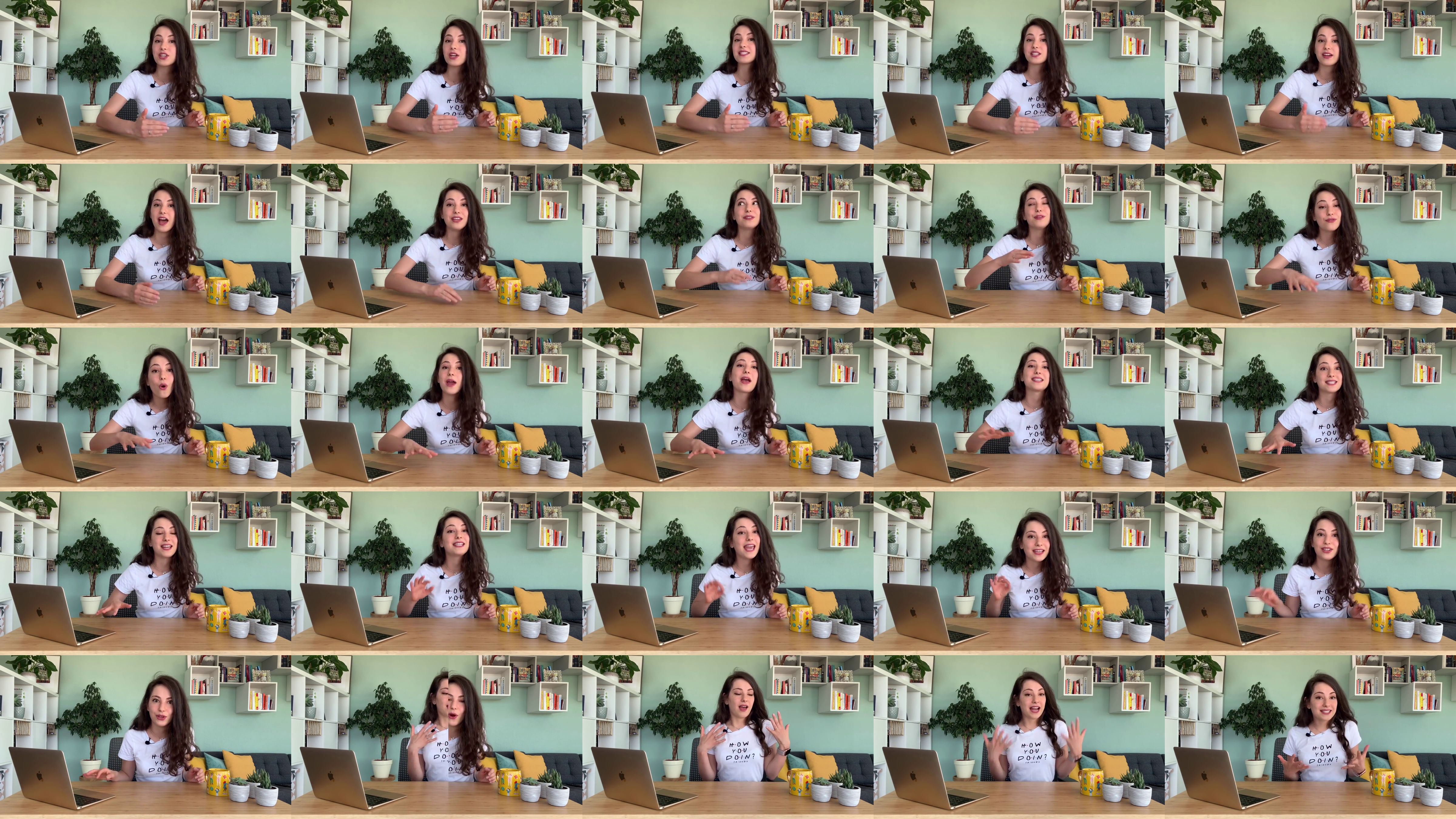}
        \caption*{\footnotesize The subject, a young woman, sits at a table... her primary movements are hand gestures}
    \end{subfigure}
    \hfill
    \begin{subfigure}[t]{0.32\linewidth}
        \includegraphics[width=\linewidth]{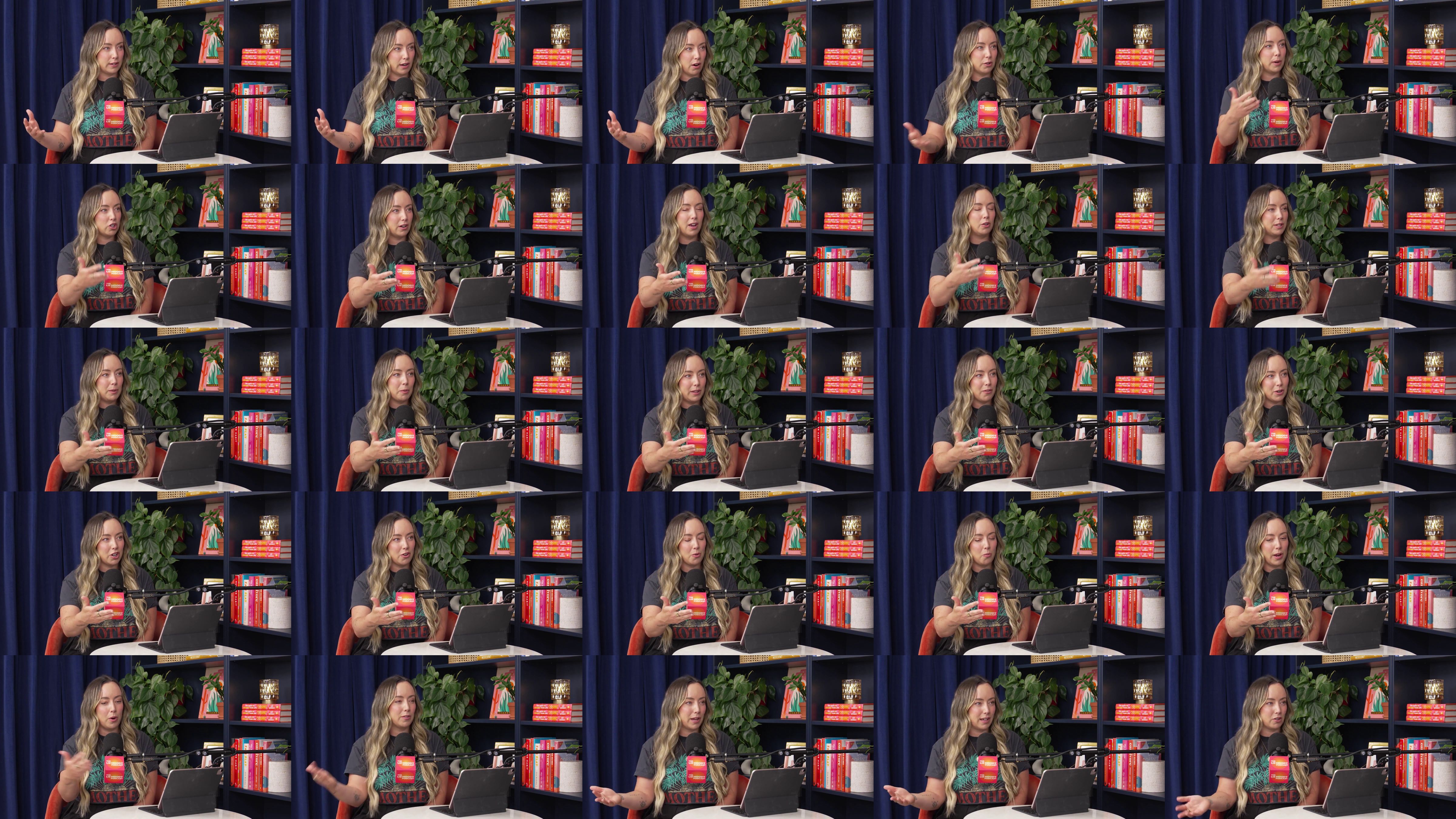}
        \caption*{\footnotesize The subject, an adult female, is seated... She engages with the camera using a combination of hand gestures... She holds a small red box and gestures with it...}
    \end{subfigure}

    \caption{\textbf{Qualitative examples from the TalkVid-Core dataset.} Each example displays sampled frames from a video clip, paired with its corresponding descriptive caption generated by Gemini 1.5 Pro. For brevity, captions are truncated.}
\end{figure*}

\begin{figure*}[htbp]
    \centering
    \captionsetup{justification=centering}
    \includegraphics[width=0.85\textwidth]{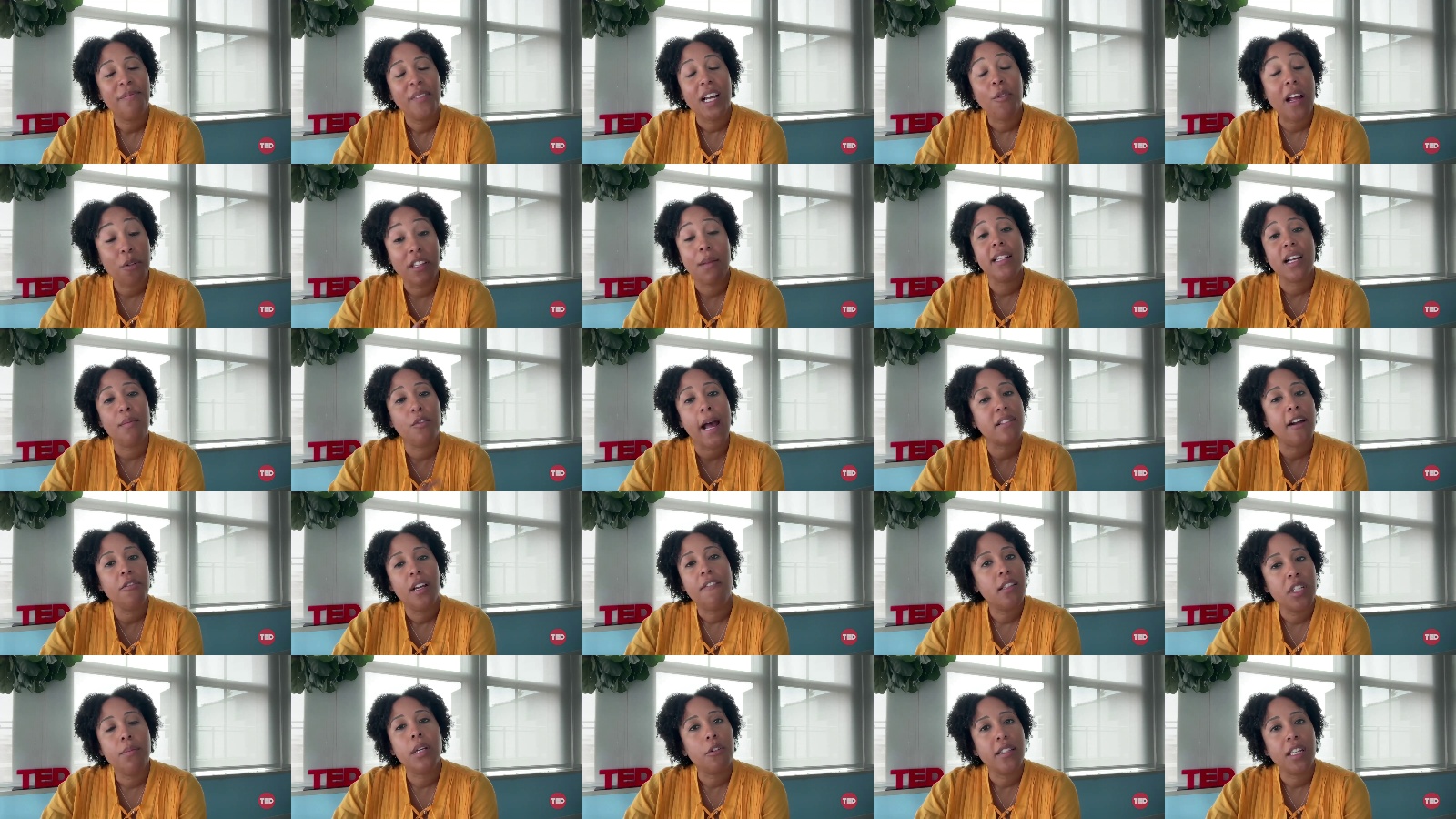}
    \caption{\textbf{Positive Example (Passes All Filters)}. This visual guideline shows an ideal case that meets all quality criteria. The subject is consistently front-facing, stable, well-lit, and occupies a significant portion of the frame. This type of clip should be labeled as \textbf{positive}.}
    \label{fig:positive_example_guideline}
\end{figure*}

\begin{figure*}[htbp]
    \centering
    \captionsetup{justification=centering}

    \begin{subfigure}[b]{0.48\textwidth}
        \includegraphics[width=0.95\textwidth]{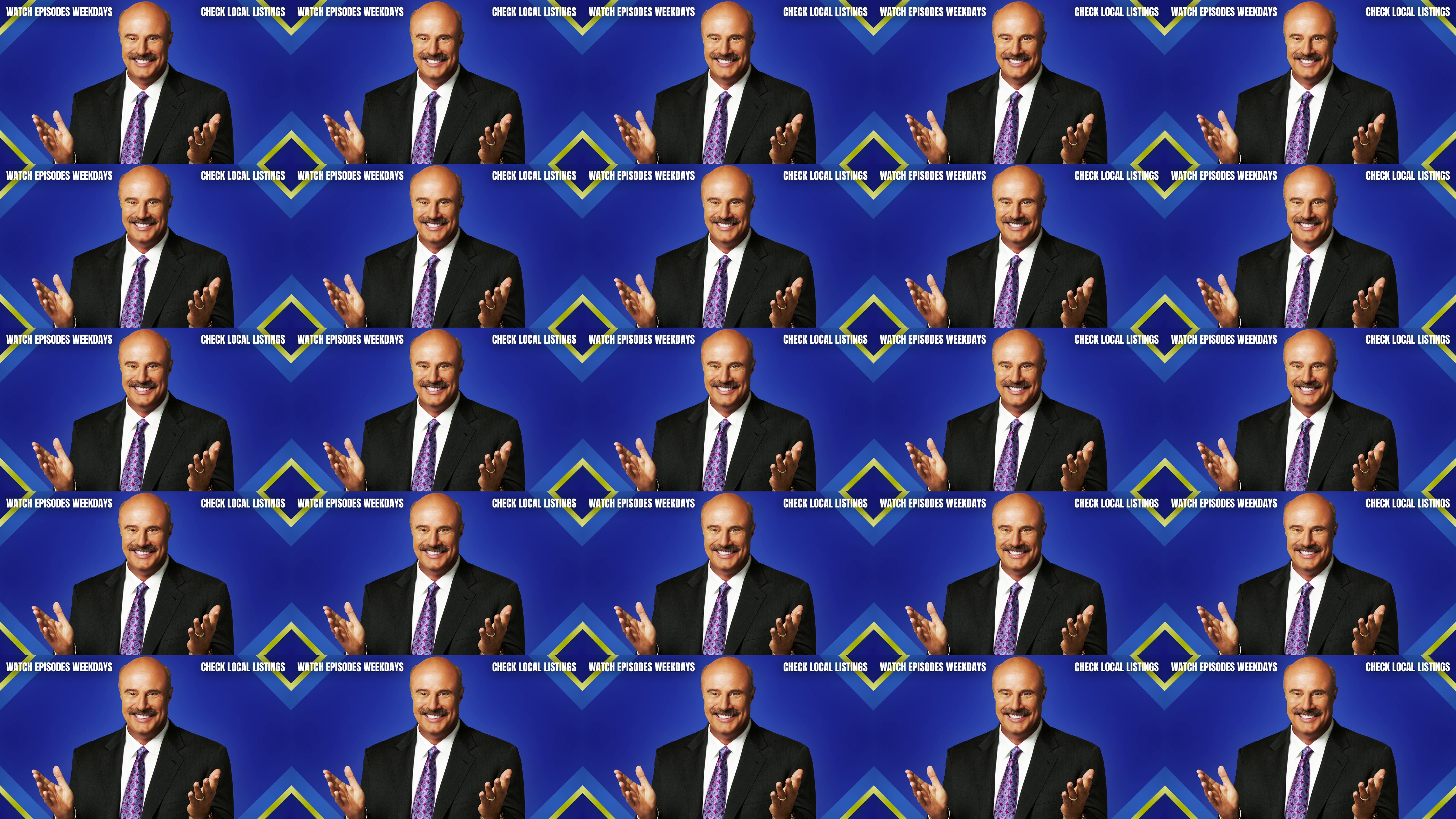}
        \caption{Fails the \textbf{Cotracker Filter} due to it remains still.}
        \label{fig:ex_neg_cotracker}
    \end{subfigure}
    \hfill 
    \begin{subfigure}[b]{0.48\textwidth}
        \includegraphics[width=0.95\textwidth]{images/negative_headmovement.jpg}
        \caption{Fails the \textbf{Head Movement Filter} due to abrupt scene changes.}
        \label{fig:ex_neg_headmovement}
    \end{subfigure}
    
    \begin{subfigure}[b]{0.48\textwidth}
        \includegraphics[width=0.95\textwidth]{images/negative_headrotation.jpg}
        \caption{Fails the \textbf{Head Rotation Filter} due to distinct head turns.}
        \label{fig:ex_neg_headrotation}
    \end{subfigure}
    \hfill
    \begin{subfigure}[b]{0.48\textwidth}
        \includegraphics[width=0.95\textwidth]{images/negative_headorientation.jpg}
        \caption{Fails the \textbf{Head Orientation Filter} as the face is not front-facing.}
        \label{fig:ex_neg_headorientation}
    \end{subfigure}

    \begin{subfigure}[b]{0.48\textwidth}
        \includegraphics[width=0.95\textwidth]{images/negative_headcompleteness.jpg}
        \caption{Fails the \textbf{Head Completeness Filter} due to facial occlusion.}
        \label{fig:ex_neg_headcompleteness}
    \end{subfigure}
    \hfill
    \begin{subfigure}[b]{0.48\textwidth}
        \includegraphics[width=0.95\textwidth]{images/negative_headresolution.jpg}
        \caption{Fails the \textbf{Head Resolution Filter} as the face is too small.}
        \label{fig:ex_neg_headresolution}
    \end{subfigure}

    \begin{center}
        \begin{subfigure}[b]{0.48\textwidth}
            \includegraphics[width=0.95\textwidth]{images/negative_dover.jpg}
            \caption{Fails the \textbf{Dover Filter} due to motion blur and low quality.}
            \label{fig:ex_neg_dover}
        \end{subfigure}
    \end{center}
    
    \caption{Visual guidelines illustrating \textbf{negative examples} for all seven filtering stages. Each sub-figure demonstrates a specific failure criterion, instructing evaluators to label such clips as \textbf{negative}.}
    \label{fig:negative_examples_guideline}
\end{figure*}


\end{document}